\documentclass[10pt, conference, compsocconf]{IEEEtran}
\usepackage{lipsum}% Just for this example
\usepackage{graphicx}
\usepackage{subfigure}
\usepackage{booktabs, multirow} % for borders and merged ranges
\usepackage{amsmath,bm}
\usepackage[final]{listings}
\usepackage[noend]{algpseudocode}
\usepackage{makecell}
\usepackage{multicol}
\usepackage{amssymb}
\usepackage{bbding}
\usepackage{hyperref}
\usepackage{balance}
\usepackage{algorithm,algpseudocode}
\algnewcommand{\To}{\textbf{To }}
\algnewcommand\Input{\item[\textbf{Input:}]}%
\algnewcommand\Output{\item[\textbf{Output:}]}%
\newcommand{\sysName}{KG-WISE}

\newcommand{\TOSA}{KG-TOSA}

\usepackage[hyphenbreaks]{breakurl}

\newcommand{\nsstitle}[1]{\noindent\textup{\textbf{#1}}}
\def\shorten{\looseness=-1} 

\usepackage{tikz}

\algnewcommand\algorithmicworker{\textbf{Worker}}
\algnewcommand\Worker{\algorithmicworker{} }

\usepackage[many]{tcolorbox}    	% for COLORED BOXES (tikz and xcolor included)
\usepackage{setspace}               % for LINE SPACING
\definecolor{black}{HTML}{000000}  
\definecolor{blue_sub}{HTML}{cde4ff}     % setting sub color to be used  
\tcbset{
    sharp corners,
    colback = white,
    before skip = 0.2cm,    % add extra space before the box
    after skip = 0.5cm      % add extra space after the box
}  
\newtcolorbox{boxH}{
    colback = sub, 
    colframe = main, 
    boxrule = 0pt, 
    leftrule = 6pt % left rule weight
}
\newtcolorbox{boxE}{
    rounded corners,
    arc = 3pt,
    colframe = black!50, 
    enhanced, % for a fancier setting,
    boxrule =0.2mm, % clearing the default rule
    borderline = {0.2mm}{0pt}{black!20}, % outer line
    left=1mm, % Left margin inside the box
    right=1mm, % Right margin inside the box
    top=1mm, % Top margin inside the box
    bottom=1mm % Bottom margin inside the box
}

\AtBeginDocument{%
  }

\begin{document}

% \title{KG-WISE: \textbf{W}ise \textbf{I}nference through \textbf{S}elective \textbf{E}mbeddings}

\title{An LLM-Guided Query-Aware Inference System for GNN Models on Large Knowledge Graphs}

\author{

\IEEEauthorblockN{Waleed Afandi}
\IEEEauthorblockA{\textit{Concordia University}
% \\
% waleed.afandi\\@mail.concordia.ca
}
\and
\IEEEauthorblockN{Hussein Abdallah}
\IEEEauthorblockA{\textit{Concordia University}
% \\
% hussein.abdallah\\@mail.concordia.ca
}
\and
\IEEEauthorblockN{Ashraf Aboulnaga}
\IEEEauthorblockA{\textit{University of Texas at Arlington}
% \\
% ashraf.aboulnaga\\@uta.edu
}
\and
\IEEEauthorblockN{Essam Mansour}
\IEEEauthorblockA{\textit{Concordia University}
% \\
% essam.mansour\\@mail.concordia.ca
}
}

% \author{Hussein Abdallah}
% \affiliation{%
%   \institution{Concordia University}
%   \country{Canada}
% }
% \email{hussein.abdallah@mail.concordia.ca}

% \author{Essam Mansour}
% \affiliation{%
%   \institution{Concordia University}
%   \country{Canada}
% }
% \email{essam.mansour@concordia.ca}
\setcounter{footnote}{1}
 \maketitle
\begin{abstract}
\sloppy
Efficient inference for graph neural networks (GNNs) on large knowledge graphs (KGs) is essential for many real-world applications. 
GNN inference queries are computationally expensive and vary in complexity, as each involves a different number of target nodes linked to subgraphs of diverse densities and structures. Existing acceleration methods, such as pruning, quantization, and knowledge distillation, instantiate smaller models but do not adapt them to the \textit{structure or semantics} of individual queries. They also store models as monolithic files that must be fully loaded, and miss the opportunity to retrieve only the neighboring nodes and corresponding model components that are semantically relevant to the target nodes. These limitations lead to excessive data loading and redundant computation on large KGs.
This paper presents {\sysName}, 
% a scalable system for adaptive and efficient GNN inference on large KGs. 
a task-driven inference paradigm for large KGs. {\sysName} decomposes trained GNN models into fine-grained components that can be partially loaded based on the structure of the queried subgraph. It employs large language models (LLMs) to generate reusable query templates that extract semantically relevant subgraphs for each task, enabling query-aware and compact model instantiation. We evaluate {\sysName} on six large KGs with up to 42 million nodes and 166 million edges. 
{\sysName} achieves up to 28$\times$ faster inference and 98\% lower memory usage than state-of-the-art systems while maintaining or improving accuracy across both commercial and open-weight LLMs.
\end{abstract}

\maketitle
\section{Introduction}
\label{sec:intro}

\begin{figure}[t]
  \vspace*{-2ex}
  \centering
  \includegraphics[width=\columnwidth]{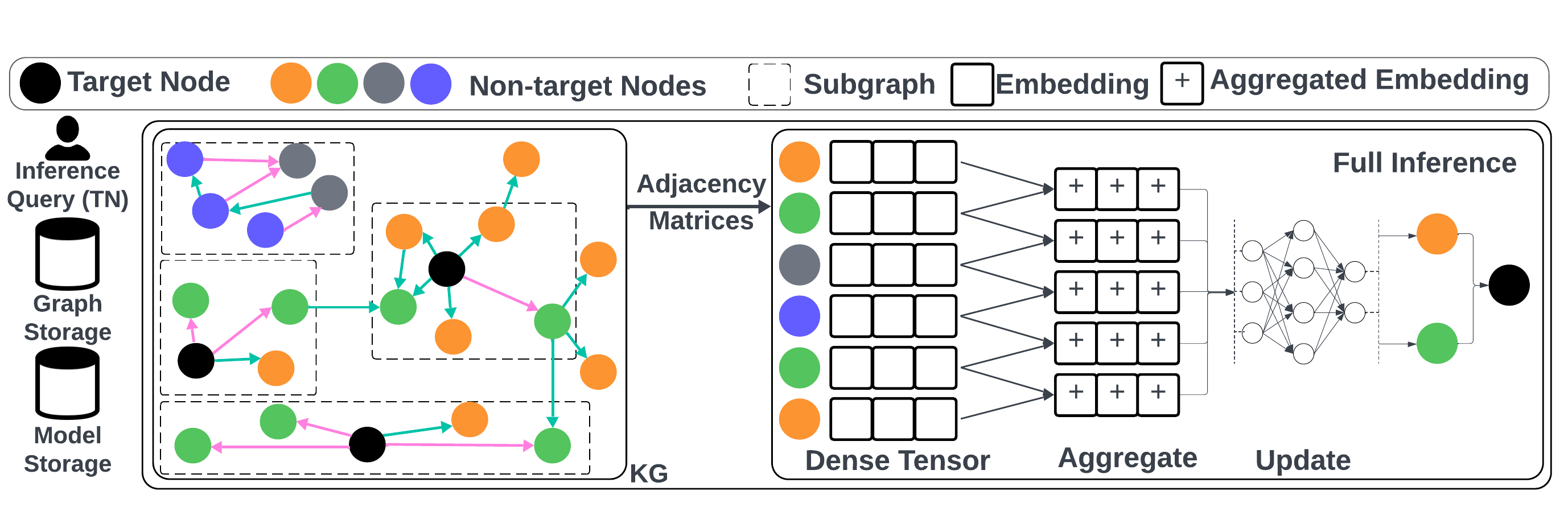}
  \vspace*{-5ex}
  \caption{A GNN inference query for target nodes ($TN$) loads the KG’s adjacency, model, and embedding matrices from storage, then performs message passing and embedding updates to produce predictions. This process is resource-intensive and scales poorly with the size of these matrices in KGs.}

  \label{figures:GNN_InferencePipeline}
  \vspace*{-4ex}
\end{figure}

Graph neural networks (GNNs) on knowledge graphs (KGs) have proven effective in many applications~\cite{GNN_KG_Survey_2022}, such as recommendation~\cite{gnnRS}, drug discovery~\cite{kgnn}, anomaly detection~\cite{GNNAnomaly}, and fraud detection~\cite{fraudDet}. 
An {\em inference query} on a KG uses a trained GNN model to predict missing information (e.g., link prediction or node classification) for a set of {\em target nodes} ($TN$)~\cite{Serverless_GNN}. 
Efficient inference is essential for deploying GNNs on large KGs. The cost depends on the density and structure of the subgraphs around the target nodes. During inference, the system loads large data from disk to main memory and GPU memory, including the entire KG adjacency, model parameters, and node embeddings. It then performs dense tensor aggregation and message passing for each node, as shown in Figure~\ref{figures:GNN_InferencePipeline}. 
This process is computationally expensive and scales poorly as the KG grows.
\textbf{This raises a central research problem: how can we perform scalable and adaptive GNN inference that tailors computation and data loading to the structure and semantics of each inference query on a large KG?}

Prior work on GNN inference acceleration focuses on converting trained models into smaller or faster versions for deployment~\cite{Inference_Accelaration_Survey_2024}. Typical methods include pruning~\cite{GCNP}, quantization~\cite{Degree-Quant}, and knowledge distillation~\cite{GKD}. These approaches mainly reduce model weights but ignore the cost of handling precomputed embeddings for non-target nodes, which must still be loaded during inference. Most systems store trained GNNs as monolithic files, forcing full model loading even when only a small part is relevant to the query. Another line of work performs inference on fixed $L$-hop neighborhoods around each target node~\cite{Serverless_GNN}. 
While this reduces computation, it treats all queries uniformly and overlooks semantic and structural differences among them. Hence, these methods fail to adapt inference to the query context and often load less relevant neighbors, leading to redundant computation on large heterogeneous KGs.

As a result, existing accelerators work well on small graphs but scale poorly in realistic KG settings. They miss the opportunity to adapt computation and data loading to each query by instantiating compact and query-specific models. Achieving this goal requires a system that performs scalable and adaptive inference tailored to the structure and semantics of each query.

GNN {\em training} accelerators improve scalability by sampling smaller subgraphs for mini-batch training. Methods such as random walk sampling (GraphSAGE~\cite{Graph-SAGE}), node-importance sampling (ShaDowGNN~\cite{Shadow-GNN} and IBMB~\cite{ibmb}), and task-oriented sampling (KG-TOSA~\cite{KGTOSA}) reduce training time and memory usage. However, these techniques are designed for training, not inference. At inference time, the full model and its complete training graph must still be loaded, regardless of the query’s size or structure. They also assume fixed neighborhoods and ignore variation in query semantics and subgraph density. Thus, they fail to tailor computation and data loading to the structure and \textit{semantics} of the GNN task on a large KG. For instance, KG-TOSA misses this opportunity by using fixed graph patterns that retrieve neighboring nodes without considering their semantic relevance to the target task.

% This paper presents {\sysName}, a
% query-aware inference system that addresses the limitations of existing inference approaches on large KGs by introducing several innovations.
% First, {\sysName} instantiates a compact query-specific inference model by adapting the trained GNN model to the structure and semantics of each \textcolor{blue}{inference task,i.e., predicting the place country. LLMs are good in-context graph learners~\cite{LLM-Graph-Learner,LLM-Temporal-Graph-Learner} that can learn the graph structure and predict missing/future nodes/edges. Thus, {\sysName} uses an LLM to identify entities and predicates in the KG schema that are semantically relevant to the inference task.
% It generates a SPARQL query to extract a subgraph from the KG consisting of these semantically relevant triples centered on the target nodes of the query.}
% This subgraph is orders of magnitude smaller than the full KG, often containing thousands of nodes compared to millions in the full graph.
% Second, {\sysName} only loads the components of the GNN model associated with this subgraph, which reduces memory and computation costs compared to loading the full model.
% To support this partial loading, {\sysName} decomposes trained GNNs into modular components, such as weights, intermediate parameters, and node embeddings, and stores them in a key-value store.
% Third, based on the subgraph, {\sysName} selects between sparse or dense tensor aggregation for efficiency. These innovations combine to significantly reduce inference time and memory usage.

This paper presents {\sysName}, a query-aware inference system that overcomes the limitations of existing GNN inference approaches on large KGs. 
{\sysName} adapts the trained GNN model to the structure and semantics of each task through three key steps.
%
% First, during training, \highlightedReplyIII{ {\sysName} uses an LLM-based method to analyze the description of the GNN task and its target nodes ($TN$). 
% The LLM identifies entities and predicates in the KG schema that are semantically related to the task and generates a SPARQL query template to retrieve the corresponding subgraph.
% This subgraph, which is much smaller than the full KG, is then used to train the model for that task. After training, {\sysName} decomposes the model into modular components, namely weights, intermediate parameters, and node embeddings, and stores them in a key-value store for partial loading.}{R3.O3}
First, during training, {\sysName} uses an LLM-based method to analyze the GNN task description and its target nodes ($TN$).
The LLM identifies relevant entities and predicates in the KG schema and generates a SPARQL query template to retrieve the corresponding subgraph.
This subgraph, significantly smaller than the full KG, is then used to train the model.
After training, {\sysName} decomposes the model into modular components, weights and node embeddings, and stores them in a key-value store for partial loading.
Second, during inference, {\sysName} reuses the stored query template to fetch a semantic subgraph relevant to the current inference query. It then instantiates a compact query-specific model by loading only the components associated with that subgraph. This design avoids redundant data movement and computation while preserving prediction accuracy.\shorten
Third, {\sysName} dynamically selects between sparse or dense tensor aggregation based on the structure of the retrieved subgraph to further improve efficiency. Together, these techniques enable scalable, query-aware GNN inference that reduces memory and computation costs while maintaining model accuracy.

We evaluate {\sysName} on the KGBen benchmark~\cite{KGTOSA}, which includes KGs with up to 42M nodes and 166M edges across six domains (e.g., DBLP~\cite{KG_DBLP}, MAG~\cite{MAG}, YAGO4~\cite{KG_Yago4}, and WikiKG~\cite{KG_Wikidata}). Tasks cover node classification and link prediction with 100–1600 target nodes to model realistic inference workloads.
We compare against inference accelerators GCNP~\cite{GCNP}, Degree-Quant~\cite{Degree-Quant}, and GKD~\cite{GKD} (when scalable), and training methods GraphSAINT~\cite{GraphSAINT}, MorsE~\cite{MorsE}, IBMB~\cite{ibmb}, and KG-TOSA~\cite{KGTOSA}. {\sysName} matches or improves accuracy while reducing inference time by up to 28$\times$ and memory by up to 98\%. Ablations isolate the gains from LLM-guided subgraphs and partial model loading; query-size, scalability, and CPU/GPU tests show consistent improvements with low preprocessing overhead. 
% To our knowledge, this is the first unified evaluation of pruning, quantization, and distillation methods on large KGs in a serving setting. 
We also evaluated {\sysName} with both commercial (Gemini, GPT4/5) and open-weight (GPT-oss, Qwen, DeepSeek) LLMs, observing comparable accuracy and efficiency across all models. 
% This confirms that {\sysName} is not tied to a specific LLM.

In summary, the main contributions of this paper are:
\begin{itemize}
  \item The first end-to-end system for scalable GNN storage and inference on large KGs; Section~\ref{sec:architecture}.

  \item A fine-grained decomposition and storage mechanism for GNN models on KGs that enables \textit{partial model loading} with minimal preprocessing overhead; Section~\ref{sec:storage}.

  \item A query-aware inference approach that uses LLM-guided query templates to extract semantically relevant subgraphs and instantiate compact models with negligible retrieval and loading cost; Section~\ref{sec:inference}.

  \item A comprehensive evaluation on large KGs showing that {\sysName} reduces inference time by up to 28$\times$ and memory by up to 98\% while maintaining or improving accuracy. 
  The results are consistent across commercial and open-weight LLMs, demonstrating that {\sysName}'s performance is not tied to any specific model; Section~\ref{sec:expermients}.\shorten
\end{itemize}

\section{Background and Related Work}
\label{sec:background}
\begin{table*}[t]
 \vspace*{-2ex}
    \centering
    \renewcommand{\arraystretch}{1.5}
    \caption{A Comparative Analysis of Existing and Proposed Inference Acceleration Techniques.}
    \scriptsize
    % \footnotesize
    \vspace*{-2ex}
    % \resizebox{\textwidth}{!}{%
        \begin{tabular}{|c|c|c|c|c|} \hline
            & \textbf{Pruning (GCNP) \cite{GCNP}} & \textbf{Quantization (DQ) \cite{Degree-Quant}} & \textbf{Know. Distillation (GKD) \cite{GKD}} & \textbf{Ours (KG-WISE)} \\ \hline
            \textbf{Optimization Overhead} & Per layer Channel Pruning & Simulation of Quantization effect & Student Model Training & Decompose $H$ into KV Store \\ \hline
            % \textbf{Optimized Element} & $F^\prime,L^\prime$ & $F^\prime$ & $E^\prime,N^\prime$ & - \\ \hline
            \textbf{Additional Training} 
            & \(\bm{\checkmark}\)
            & \(\bm{\checkmark}\) 
            & \(\bm{\checkmark}\) 
            & \(\bm{\times}\) \\ 
            \hline
            \textbf{Storage Mechanism} & Disk based single file & Disk based single file & Disk based single file & Disk based \& Key-Value store \\ \hline
            \textbf{Partial Graph Loading} & \(\bm{\times}\) (Full Graph) & \(\bm{\times}\) (Full Graph) & \(\bm{\checkmark}\) (Randomly sparsed) & \(\bm{\checkmark}\) (Query based) \\ \hline
            \textbf{Partial Embedding Loading} & \(\bm{\times}\) (Full Graph) & \(\bm{\times}\) (Full Graph) & \(\bm{\times}\) (Full Graph) & \(\bm{\checkmark}\) (1 hop) \\ \hline
            \textbf{Forward Pass} & Full & Full & Random Sparse & Query Aware \\ \hline
            \textbf{Inference Time Complexity} & $O(L^{\prime}N^2F^{\prime} +L^{\prime}NF^{\prime 2})$ &$O(LN^2F^{\prime} +LNF^{\prime 2})$&$O(LN^{\prime 2}F +LN^{\prime}F^2)$ &$O(LN_{SG}^2F_{SG} +LN_{SG}F_{SG}^2)$ \\ \hline
            % $F^{\prime}$ is the quantized F. 
        \end{tabular}%
            \label{tab:comparison_transposed}
             \vspace*{-3ex}
\end{table*}
\subsection{Training GNNs on KGs}
Training a GNN model on a KG takes the graph structure and its node features as input.
The node features are used to initialize node embeddings. If node features are missing, node embeddings are initialized randomly. The training process generates node representations for downstream tasks along with
the model parameters. Unlike GNN methods designed for homogeneous graphs,
GNN methods for heterogeneous graphs, such as KGs, extend traditional approaches to support diverse
node and edge types. These methods utilize multi-layer architectures,
such as Relational Graph Convolutional Networks (RGCNs)~\cite{RGCN} and often
adopt sampling-based mini-batch training~\cite{GraphSAINT, Shadow-GNN, MorsE}. For
instance, RGCN layers aggregate embeddings from neighboring nodes based on
specific relation types. The hidden embedding of a node $i$ at layer $l+1$ is computed
as:
\begin{equation}
    \label{eq_rgcn}h_{i}^{(l+1)}= \sigma\left(\sum_{r \in \mathcal{R}}\sum_{j
    \in N_i^r}\frac{1}{c_{i,r}}W_{r}^{(l)}h_{j}^{(l)}+ W_{0}^{(l)}h_{i}^{(l)}\right
    )
\end{equation}

$h_{i}^{(l+1)}$ is the updated embedding of node $i$, $\sigma$ is an activation function,
$N_{i}^{r}$ is the set of neighbors of node $i$ under relation $r$, $c_{i,r}$ is
a normalization constant (e.g., $c_{i,r}= |N_{i}^{r}|$), $W_{r}^{(l)}$ is the
weight matrix for relation $r$ at layer $l$, and $W_{0}^{(l)}$ is the layer-specific
weight matrix for self-loops.
Training on large KGs is computationally expensive due to the need to aggregate messages from multi-hop neighbors, where nodes are connected through multiple edges of specific types ($\mathcal{R}$). This becomes more demanding in deep GNNs as information propagates through many layers, especially with a large $|\mathcal{R}|$~\cite{KGTOSA}.
The computational overhead increases due to the size and heterogeneity of KGs, since training iterates over all the relations $\mathcal{R}$.
The size of nodes and edges affects significantly the overall computations, as illustrated
in Equation~\ref{eq_rgcn}.\shorten

Mini-batch training scales GNNs by sampling subgraphs to reduce memory usage during training. Various RGCN-based methods employ different sampling strategies: GraphSAINT~\cite{GraphSAINT} uses random walk sampling, IBMB~\cite{ibmb} applies personalized PageRank, Shadow-GNN~\cite{Shadow-GNN} adopts node-importance sampling, MorsE~\cite{MorsE} employs structure-aware sampling, and KG-TOSA~\cite{KGTOSA} performs task-oriented sampling. 
However, these methods still require loading the entire trained model and the full graph at inference time (Figure~\ref{figures:GNN_InferencePipeline}). 
%%%Essam Retrun for CRV%%%
% {
% \color{green}
In particular, KG-TOSA relies on fixed graph patterns that retrieve neighboring nodes without considering their semantic relevance to the GNN task, resulting in unnecessary nodes and edges during both training and inference. In contrast, {\sysName} uses an LLM-guided extraction method to identify semantically relevant non-target node types, producing a more compact and efficient model during training. At inference, {\sysName} selectively loads only the necessary model components to instantiate a query-specific model on demand.

%}

\subsection{Traditional GNN Inference Pipeline}
The typical inference pipeline in GNN methods
%, as implemented in existing GNN methods,
begins with an inference query $\mathcal{IQ}$($TN$) that requests predictions (e.g.,
Node Classification (NC) or Link Prediction (LP)) for a set of target nodes in a
KG. The pipeline involves loading the entire KG and trained GNN model into memory, followed by GNN aggregation over dense node embeddings across multiple layers, as illustrated in Figure~\ref{figures:GNN_InferencePipeline} and Equation~\ref{eq_rgcn}.
The inference pipeline performs a forward pass on a dense graph, where $N$ is the number of nodes, $L$ is the number of layers, and $F$ is the average node embedding size per layer~\cite{GNN_Complexity}. The time complexity is $O(LN^{2}F + LNF^{2})$ and the space complexity is $O(N^{2}+ LF^{2}+ LNF)$. This process is memory-intensive and computationally costly, particularly for large graphs. It may also involve unnecessary computations, such as calculating embeddings for unreachable non-target nodes, which do not impact target node embeddings, as shown in Figure~\ref{figures:GNN_InferencePipeline}.

\nsstitle{Full Inference vs. Batched Inference:} In full inference, the entire graph
is used for aggregation, where each node aggregates information from all its
neighbors across the graph. Although accurate, this approach is computationally
prohibitive for large graphs. In contrast, batched inference divides target nodes into smaller batches and
processes them iteratively via subgraphs sampling. This approach reduces memory usage
but may compromise aggregation accuracy as it limits access to a node's full neighborhood.

%%%Essam Retrun for CRV%%%
% {
% \color{green}
Graph ML libraries like PyG~\cite{pytorch-geometric} and DGL~\cite{DistDGL} support both training and inference and continue to adopt new techniques. However, unlike {\sysName}, they load the entire model and graph into memory without optimizing for query relevance or avoiding unnecessary components. Most of the baseline implementations used in our study are available through PyG.
%}

\subsection{GNN Inference Acceleration Methods}

Given a trained GNN model $M$, inference acceleration aims to derive a smaller and faster model $\widetilde{M}$ that preserves the accuracy of $M$. These accelerators typically require access to the graph and often involve retraining. The instantiated model $\widetilde{M}$ may change architecture, parameter precision, or weight representations. Existing techniques fall into three main categories: pruning, quantization, and knowledge distillation. Table~\ref{tab:comparison_transposed} compares these methods with {\sysName}.

\nsstitle{Homogeneous vs. Heterogeneous Graphs.}
Most existing accelerators are designed for homogeneous graphs, where all nodes share the same type and embeddings are generated on demand during inference. In heterogeneous graphs such as KGs, however, the model stores large precomputed embeddings for non-target nodes to capture semantic diversity. These embeddings often dominate model size and must still be fully loaded at inference time, even when only a small subset is relevant to the query. In the following, we review the main categories of GNN inference acceleration methods and discuss their general principles and limitations.

\nsstitle{GNN Pruning}: This approach aims to identify and prune specific weights
$W$ in the GNN model $M$ without compromising its accuracy~\cite{GCNP,GEBT}. The
approach applies an additional step after the model $M$ has been trained and involves
per-layer channel pruning. This approach necessitates extra training to identify
the weights to remove~\cite{GCNP,GEBT}. The pruning reduces the model size by reducing
the complexity of layer $L^{\prime}$ via its reduced hidden layer feature size $F
^{\prime}$, which contributes to a smaller $W$. Instantiating the inference pipeline
requires loading both the full model $\widetilde{M}$ and the complete graph. The pruned model $\widetilde{M}$
is stored on disk as a single file.
In this approach, the inference process performs computation on the entire graph, regardless of the target node set, which is expensive and which {\sysName} aims to avoid.

\nsstitle{GNN Quantization}: The quantization approach reduces the numerical
precision of the model parameters. It represents the hidden layer features with fewer
bits, yielding an $F^{\prime}$ that contributes to faster matrix multiplications
of $W$ and a smaller version of the model, $\widetilde{M}$, while maintaining performance
comparable to $M$~\cite{SGQuant,Degree-Quant}. Like pruning, quantization requires
additional training to instantiate $\widetilde{M}$. This training employs a
Quantization Aware Training (QAT) framework, where the model $M$ is trained as if
its weights $W$ are quantized. This prepares the model to handle the effects of
quantization in its final deployment stage and adapt to quantization effects to
mitigate accuracy loss. Quantization reduces model size by using smaller weights
$W$, but it does not prune parameters or layers. Both $M$ and $\widetilde{M}$ models
are stored as a single file on disk. As before, inference requires full loading of
both the graph and model $\widetilde{M}$.\shorten

\nsstitle{GNN Knowledge Distillation (KD)}: This approach focuses on encoding and
transferring geometric information graph structure from a
teacher GNN model to a student GNN model. The student model is designed to be smaller,
with fewer $N^{\prime}$ nodes and $E^{\prime}$ edges to achieve faster inference
and comparable model performance to the teacher~\cite{tinyGNN,GLNN}. For instance,
methods, such as GKD~\cite{GKD}, train a teacher model first and then train a
student model to replicate the teacher's geometric understanding. The student model
is trained using a randomly sparse graph, which is generated by removing nodes
and edges randomly from the KG.
%%%Essam Retrun for CRV%%%
% \textcolor{green}{
This knowledge transfer enables the student model
to match the teacher's performance while being more resource-efficient. However,
during inference, these methods still require loading the full model and graph into memory.\shorten
%}

\noindent\textbf{Limitations of Existing Accelerators.}
Across pruning, quantization, and KD, the instantiated model $\widetilde{M}$ is always a \textit{single static artifact} stored as one file. None of these methods decompose embeddings by node type or support partial loading. As a result, even inference over a small number of target nodes, common in real-world deployments~\cite{RAIN,Serverless_GNN}, forces full-model materialization, leading to high memory overhead and inefficient inference on KGs.

{\sysName} differs from traditional black-box accelerators that operate on arbitrary pre-trained models. It follows a task-driven inference paradigm where training and inference are aligned through semantic subgraph extraction. Only task-relevant model components and embeddings are loaded into memory. This goes beyond standard sampling-based pipelines and enables more efficient inference.

% These limitations motivate our query-aware design, where model footprint scales with the semantic scope of the inference query instead of the size of the entire KG.

\section{{\sysName} System Overview}
\label{sec:architecture}

\begin{figure}[t]
 \vspace*{-2ex}
  \centering
  \includegraphics[scale=0.9]{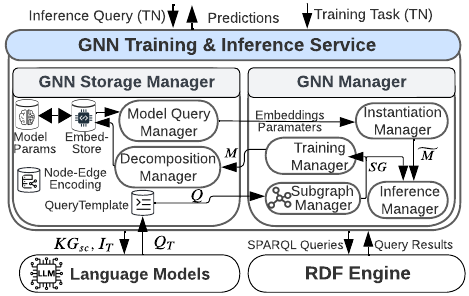}
   \vspace*{-2ex}
  \caption{
  {\sysName} orchestrates training and inference on large KGs through LLM-guided subgraph extraction, fine-grained model storage, and query-aware model instantiation.
  }  
  \label{figures:kgwise_architecture}
 \vspace*{-1ex}
\end{figure}

{\sysName} addresses the challenge of performing scalable and adaptive GNN inference that tailors computation and data loading to the structure and semantics of each query on large KGs. 
It introduces three key innovations: 
(1) an LLM-guided method that generates reusable SPARQL query templates for extracting semantically relevant subgraphs, 
(2) fine-grained model decomposition that enables partial model loading, and 
(3) query-aware model instantiation that adapts inference to each query’s subgraph structure. 
Figure~\ref{figures:kgwise_architecture} presents the architecture of {\sysName}. 
The system consists of a \textit{GNN Manager} and a \textit{GNN Storage Manager}, supported by an RDF engine for knowledge graph storage and metadata management. 
We next describe each component following the system workflow.

\nsstitle{LLM-guided Query Template and Training Phase:}
During training, {\sysName} learns a compact task-specific model from semantically relevant subgraphs instead of the full KG. A single task refers to a specific target node type, having specific related relations, and specific tails or labels. These subgraphs are retrieved via SPARQL queries, and we generate a query template for these queries,  $Q_T$, using an LLM-guided procedure (Algorithm~\ref{alg:query_gen}), \textit{executed once before training}.

The LLM is prompted with both the task description and schema statistics.
% \footnote{All prompts and query templates used by {\sysName} are in the \ref{sec:appendix} Appendix.}
All prompts and query templates used by {\sysName} are in the \ref{sec:appendix} Appendix.
These statistics include triple-type frequencies and relation coverage, which implicitly capture schema density.
 This information is retained during schema pruning (Line 2) and reused when selecting basic graph patterns (BGPs) (Lines 3--4), guiding the LLM toward structurally relevant and well-populated relations.
The verified BGPs are then compiled into a reusable query template $Q_T$ (Lines 5--7), which is applied during both training and inference to extract a dense, task-aligned subgraph without re-invoking the LLM.
Given schema-level BGPs \( \mathcal{B} = \{(s_i, p_i, o_i)\} \) and a target node type \( VT \), KG-WISE uses an LLM to generate a constrained SPARQL query template for task-relevant subgraph extraction:\shorten

\begin{equation}
\label{eq:sparql}
Q_T = g(\mathcal{B}, VT, KG_{sc})
\end{equation}

% \( \mathcal{S} \)
where $KG_{sc}$ is the KG schema and \( g(\cdot) \) is a prompt-guided function. The prompt enforces schema-valid template population, correct edge directionality, and \( VT \)-rooted nested sub-queries combined via \texttt{UNION}. Schema verification prevents hallucinated relations and ensures reproducibility.

\begin{algorithm}[!t]
% \caption{\small \textsc{LLM-Guided Subgraph Query Generation (Training-Time Only)}}
\caption{\small \textsc{LLM-Guided Subgraph Query Generation}}
\label{alg:query_gen}
\scriptsize
\begin{algorithmic}[1]
\Input Knowledge Graph $KG$, Schema $KG_{sc}$, Number of Hops $K$, Task Instruction $I_T$, SPARQL Example $Q_E$
\Output Query Template $Q_T$
\Function{Subgraph\_Query\_Gen}{$KG_{sc}, K, I_T$}
    \State $I_{SF} \gets$ suggestTaskFeatures($I_T$)
    \State $KG_{ps} \gets$ pruneSchema($KG_{sc}, K, I_T$)
    \State $I_{BGP} \gets$ mapToBGP($I_{SF}, KG_{ps}$)
    \State $I_{BGP} \gets$ verifyBGPs($I_{BGP}, KG_{ps}$)
    \State $Q_T \gets$ BGPToSPARQL($I_{BGP}, Q_E$)
    \State $Q_T \gets$ verifySPARQL($Q_T, KG_{ps}$)
    \State \Return $Q_T$
\EndFunction
\end{algorithmic}
\end{algorithm}

\nsstitle{The Training Manager} executes $Q_T$ on the RDF engine to retrieve the task subgraph $SG$ and trains a GNN model $M$ on it. $SG$ is agnostic to the underlying GNN architecture and can be readily integrated with other GNN models to improve their training and inferencing quality.
After training, the \textit{Decomposition Manager} splits $M$ into node embeddings and model parameters, stored in a key-value (KV) store for fine-grained access. 
Each embedding chunk is indexed by node type to allow selective retrieval during inference. 
The query template $Q_T$ and model metadata are stored in the RDF engine for reuse. 

\nsstitle{A Query-aware Inference Phase:}
At inference time, {\sysName} adapts computation and data loading to the semantics of each query. Given a query targeting nodes $TN$, the \textit{Subgraph Manager} retrieves the stored template $Q_T$ from the RDF engine and executes it to extract the semantic subgraph $SG$ relevant to $TN$. 
No additional LLM calls are required. 
The \textit{Instantiation Manager} collaborates with the \textit{Model Query Manager} to fetch only the model components (weights $W$ and embeddings $E(RN_M)$) needed for $SG$. 
These components are combined to instantiate a compact query-specific model $\widetilde{M}$. 
Depending on subgraph sparsity, the Instantiation Manager selects sparse or dense tensor aggregation to optimize computation. 
The \textit{Inference Manager} then performs the forward pass on  $\widetilde{M}$ to generate predictions for $TN$, achieving efficient and accurate inference with minimal memory use.

\nsstitle{Storage and Metadata Management:}
{\sysName} employs a dual-store design for efficient data access. 
An RDF engine stores the KG and model metadata, including schema, task definitions, and query templates. 
This enables semantic subgraph retrieval through SPARQL queries using built-in RDF indices. A KV-store manages numerical tensors such as embeddings and model parameters, chunked and indexed by node type for partial loading. 
This separation of semantic metadata and numerical tensors allows {\sysName} to perform both graph retrieval and fine-grained model access efficiently.

\nsstitle{Discussion:}
Unlike prior systems that load the full model and graph for each inference query, {\sysName} performs adaptive GNN inference by tailoring data loading and computation to each query’s structure and semantics. 
Its LLM-guided subgraph templates and fine-grained model decomposition enable partial model loading and compact model instantiation. 
While this mechanism also benefits training efficiency, its primary impact lies in reducing inference time and memory usage on large heterogeneous KGs.

\section{The {\sysName} Storage Mechanism}
\label{sec:storage}

This section presents {\sysName}'s model decomposition approach and its efficient storage and retrieval methods.

\subsection{GNN Model Layout and Storage Mapping}
\begin{figure}[t]
\vspace*{-4ex}
  \centering
  \includegraphics [width=\columnwidth]{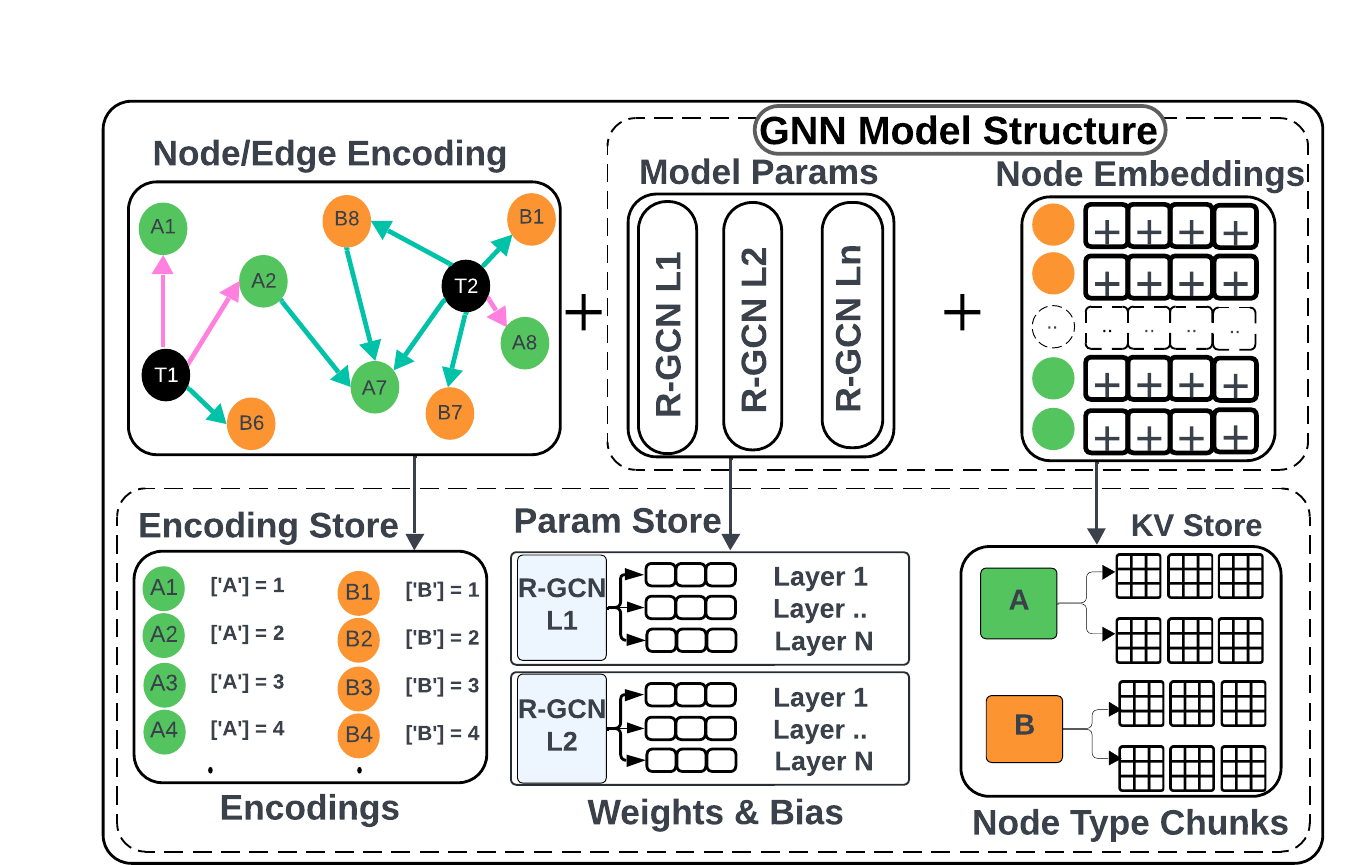}
  \vspace*{-4ex}
       \caption{The top half shows an encoded KG and a GNN trained on it. The GNN's main components are the learned parameters and \textbf{non-target node embeddings}. The bottom half illustrates how KG-WISE maps the encoding and decouples the GNN components. The node embeddings are stored in a key-value store as as row-wise chunks grouped by node type in a Zarr KV store.\shorten}

 \label{fig:GNNLayout}  
  % \vspace*{-2ex}
\end{figure}

ALL GNN models have a typical layout, shown in the top half of Figure~\ref{fig:GNNLayout}, consisting of three core components: node/edge encodings, model parameters, and node embeddings, all linked to an encoded KG. Node/edge encodings are unique identifiers for graph elements. They help transform the inference-related subgraph into adjacency matrices that align the KG with the model. Each node has a unique identifier that remains consistent during both training and inference to ensure stable mapping. Each layer's weights  are stored in the Parameter Store. Node embeddings capture the graph semantics of non-target nodes. These embeddings are learned during training and are stored as $m \times n$ arrays, where $m$ is the number of nodes and $n$ is the embedding size. Excessively increasing $n$ leads to high memory usage.\shorten

{\sysName} decomposes the GNN model and stores its components in a scalable storage system, as shown in Figure~\ref{fig:GNNLayout}. The Encoding Store maps graph elements, such as nodes and edges, to their respective encodings to enable efficient adjacency matrix construction. The Parameter Store manages the model’s weights  to ensure quick access during inference. Node embeddings are divided into chunks by node type, and these chunks are stored in a KV store, indexed by node type and node ID for efficient retrieval. This modular storage design supports random access to relevant data during inference, which helps reduce data loading and memory usage.\shorten

\subsection{The Decomposition Algorithm} 
Algorithm~\ref{alg:GNN_Decomposition} decomposes the trained GNN model and stores its components in a KV store. It first iterates over all node types, retrieves their embeddings (Lines~2--3), and chunks them by size and embedding dimension before writing them to disk (Line~4). As node encoding is already handled during training, no additional processing is needed when saving embeddings. Chunking divides large embeddings into smaller, manageable parts, improving memory efficiency and retrieval speed. For relation-specific weights, the algorithm traverses each model layer (Lines~5--6), extracts the corresponding weights (Line~7), and writes them to disk (Line~8). These relational weights are stored together as a single file for each layer.

The node embeddings are chunked and stored in a KV store, while the model parameters and weights are saved separately in the Parameter Store. The model parameters are much smaller than the embeddings. Hence, storing them separately is more efficient. This separation allows for independent and efficient access to node embeddings and parameters. {\sysName} supports parallel storage and retrieval operations, which is useful for large GNN models. As the graph size and embedding dimensions increase, the KV store provides scalable storage, which ensures fast processing and efficient retrieval. The decomposition algorithm integrates seamlessly with the KV store.\shorten

\begin{algorithm}[t]
\caption{\textsc{GNN Model Decomposition}}
\label{alg:GNN_Decomposition}
\scriptsize
\begin{algorithmic}[1]
    \Input \textit{Model $M$; Chunk size $c$; Embedding dim $e$; Key-Value Store $KVS$}
    \Output Decomposed Model
    \Function{Decompose}{$M$, $c$, $e$, $KVS$}
        \For {each vertex type $v_{t} \in M_{vertexTypes}$}
            \State $E \gets M_{v_{t}}^{emb}$ \Comment{Retrieve embeddings for vertex type $v_t$}
            \State $\text{KVS.write}(E, c, e)$ \Comment{Store embeddings in KVS in chunks}
            
        \EndFor
        \For {each layer $l \in M_{layers}$}
            \For {each relation type $r_{t} \in M_{l}$}
                \State $W_{l,r} \gets {Weights}(l,r)$
                \State $\text{storeConvWeightsToDisc}(W_{l,r})$ \Comment{Store layer weights to disc}
            \EndFor
        \EndFor
        
        % \State $\text{storeConvLayersToDisc}(M)$ \Comment{Store $M$ Conv-Layers to disc}
    \EndFunction
\end{algorithmic}
\end{algorithm}

\subsection{Efficient Storage and Retrieval}

\nsstitle{Efficient Chunking and Indexing of Node Embeddings}
{\sysName} uses chunking to reduce I/O overhead during inference by loading only the embeddings of nodes that actually appear in the inference subgraph. Chunking is \emph{row-wise}: each chunk contains full embedding vectors for a subset of nodes, not parts of
%partial dimensions of 
a single embedding vector. This enables KG-WISE to retrieve embeddings for relevant non-target nodes by type and ID, rather than loading the entire embedding matrix. The chunk size is configurable and selected based on graph size and available memory. We discuss this trade-off and provide empirical
guidance in the \ref{sec:appendix} Appendix.

\nsstitle{Zarr KV Store}
We use Zarr~\cite{Zarr} as the underlying KV store because it provides built-in block indexing over chunked arrays. Each node type is stored as a separate Zarr array, internally indexed by contiguous embedding blocks. During inference, the Model Query Manager fetches only the chunks containing the node IDs present in the extracted subgraph. This design differs from approximate nearest neighbor (ANN) indices \cite{faiss, scann}, as {\sysName} retrieves exact embeddings by node ID rather than similar embeddings by vector similarity search. Zarr’s chunk-level indexing ensures fast access without scanning unrelated embeddings. Each GNN task maintains its own Zarr instance, and updates are rare since embeddings only change when new neighbor nodes are introduced.We chose to use Zarr as our underlying storage format due to its superior support for multi-dimensional chunking and high-performance compression (e.g., Blosc). While PyG’s standard \textit{FeatureStore} provides a standardized API for GNN data access, it is fundamentally an abstraction layer that requires a storage backend. This organization ensures scalable storage, minimizes unnecessary loading, and supports efficient partial retrieval during query-aware inference.

\section{The {\sysName} Query-Aware Inference Pipeline}
\label{sec:inference}

\begin{figure}[t]
\vspace*{-2ex}
  \centering
  \includegraphics [width=\columnwidth]{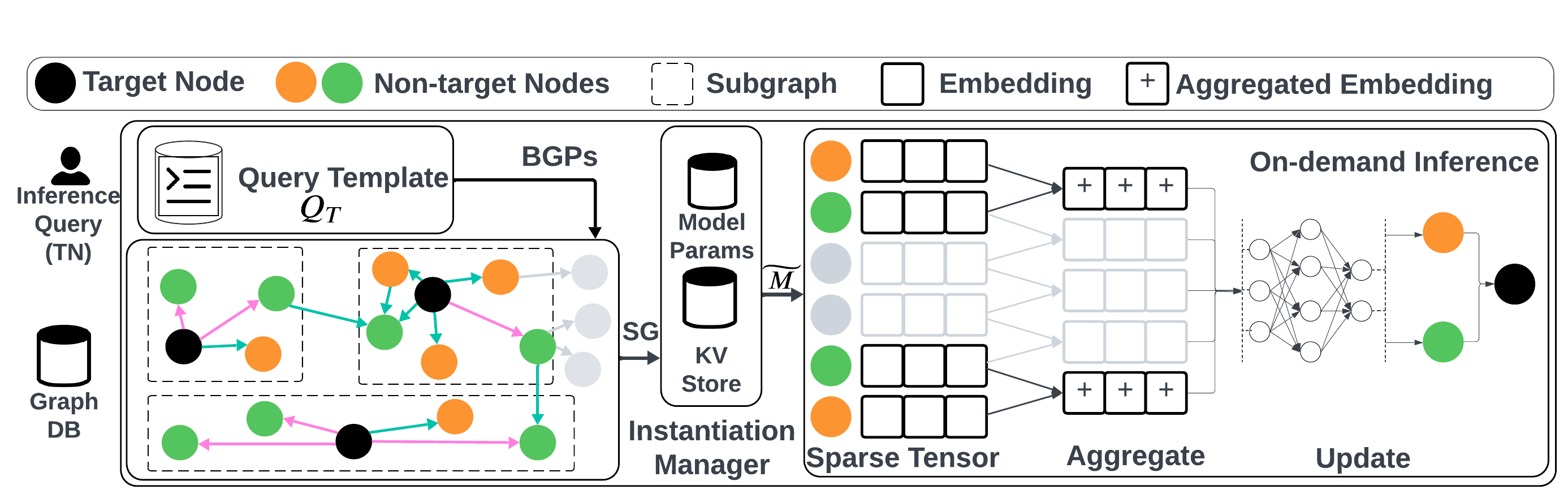}
  \vspace*{-3ex}
   \caption{ {\sysName} inference pipeline. Given an inference query, {\sysName} loads a stored SPARQL template to extract a semantically relevant subgraph $SG$, then instantiates a compact model $\widetilde{M}$ by loading only the required embeddings and weights from the KV store. Inference is executed on-demand using sparse tensor aggregation over $SG$, avoiding full model loading.}
  \label{figures:KGWISE_InferencePipeline}  
  % \vspace*{-1ex}
\end{figure}

This section presents our query-aware inference pipeline, which tailors computation and data loading to each query’s structure and semantics. {\sysName} uses an LLM-generated SPARQL template (created once during training) to extract a relevant subgraph and instantiate a compact, query-specific model $\widetilde{M}$. As shown in Figure~\ref{figures:KGWISE_InferencePipeline}, only the necessary embeddings are loaded, enabling much smaller model instances. As a result, {\sysName} achieves up to 28$\times$ faster inference and 98\% lower memory usage while maintaining comparable or better accuracy.\shorten

\subsection{Extracting the Inference-related Subgraph}
During inference, GNNs aggregate embeddings from the neighbors of each target node $TN$. In large KGs, many neighbors are unreachable or semantically irrelevant, and loading their embeddings adds computation and memory cost without improving accuracy. {\sysName} avoids this overhead by extracting a compact inference subgraph $SG$ that contains only semantically relevant nodes within $K$ hops of $TN$.
The RDF engine executes a SPARQL query derived from the stored task template $Q_T$ (Algorithm~\ref{alg:infer_subgraph_llm}). Since the template is generated once during training, no LLM calls are needed at inference time. This ensures consistency across repeated queries and avoids runtime prompt overhead.

The retrieved triples are then deduplicated and encoded. Node, relation, and label encodings are preserved to match those used during training, ensuring a seamless mapping between $SG$ and the stored embeddings. The final $SG$ is exported as a list of triples and converted to a PyG-compatible format with the same node identifiers and types used during model training.
Algorithm~\ref{alg:infer_subgraph_llm} shows this process. In Line 2, the template $Q_T$ is instantiated with the target nodes $TN$. In Line 3, the concrete SPARQL query is executed to collect the triples. In Line 4, duplicates are removed. Lines 5--6 apply node and relation encodings, and Line 7 attaches label encodings. The fully encoded inference subgraph $SG$ is returned in Line 8. This produces a dense, semantically filtered subgraph that enables partial model loading during inference.

\begin{algorithm}[!t]
\caption{\textsc{Inference Subgraph Extraction}}
\label{alg:infer_subgraph_llm}
\scriptsize
\begin{algorithmic}[1]
\Input Query Template $Q_{T}$, Knoweldge Graph $KG$, Target Nodes $TN$, SPARQL Endpoint $SP$
\Output Inference Subgraph $SG$
\Function{Inference\_SG\_Extractor}{$Q_T$, $TN$, $KG$, $SP$}
    \State $Q \gets$ instantiateQuery($Q_{T}$, $TN$)
    \State $Triples \gets$ executeSPARQL($KG$, $SP$, $Q$)
    \State $SG' \gets$ dropDuplicates($Triples$)
    \State $SG' \gets$ nodeEncodings($SG', KG$)
    \State $SG' \gets$ relationEncodings($SG', KG$)
    \State $SG \gets$ labelEncodings($SG', KG$)
    \State \Return $SG$
\EndFunction
\end{algorithmic}
\end{algorithm}

% \begin{algorithm}[t]
% \caption{{\sysName} Model Construction}
% 	\label{alg:GNN_Construction}
%         % \algsetup{linenosize=\small}
%         \small
% 	\begin{algorithmic}[1]
%              \Input \textit{Inference Sub-Graph (SG); Model File-Store $MFS$;Model Definition $M_{D}$; Key-Value Store $KVS$}
%     	\Output On-demand Model $\widetilde{M}$.
%     	\Function{Construct}{SG, $M_D$, $KVS$}
%                 \State $ \widetilde{M} \leftarrow Init(M_{D})$
%                 \State $ \widetilde{M}_{ConvLayers} \leftarrow MFS.loadConvLayers(M_{D})$
%         	\State $\mathcal{V}  \leftarrow      
%                 \textrm{getVertices( }SG )$  
%                 \For {${v} \in \mathcal{V} $}
%                    \State $v_{type} \leftarrow Type(v)$
%                    \State $VT_{dic} [v_{type}].append(v)  $ %{\mathcal{V}}
%                 \EndFor
        	
%         	\For {$V_{Type} , V_{Idx} \in VT_{dic} $}
%         		% \State $u \leftarrow v$
%                    \State $VT_{{rows}} \leftarrow \text{KVS.load}(V_{Type}).{Rows} $
%                    \State $VT_{{cols}} \leftarrow \text{KVS.load}(V_{Type}).{Cols} $
%                    \State $VT_{{emb}} \leftarrow \text{KVS.load}(V_{Type}, V_{Idx}).{Emb} $
%                    \State $\widetilde{M}_{VT} \leftarrow \text{SparseTensor}(VT_{{rows}}, VT_{{cols}}, VT_{{emb}} )  $
%                    %\State $  M^\prime _{V_{T}} \leftarrow  VT_{sparse}$
                                      
%           	\EndFor
%         	%\State $ \widetilde{M}_{ConvLayers} \leftarrow M_{ConvLayers}$
%     	\EndFunction
% 	\end{algorithmic} 
% \end{algorithm} 
\begin{algorithm}[t]
\caption{\textsc{{\sysName} Model Instantiation }}
\label{alg:GNN_Construction}
\scriptsize
\begin{algorithmic}[1]
    \Input \textit{Inference Subgraph (SG); Model File-Store (MFS); Model Definition $M_{D}$; Key-Value Store (KVS)}
    \Output On-demand Model $\widetilde{M}$.    
    \Function{Construct}{$G_{IR}$, $M_D$, $KVS$}
     \State $\widetilde{M} \gets Init(M_{D})$ \Comment{Initialize model structure based on $M_D$}
        \State $\widetilde{M}.loadConvLayers(MFS)$ \Comment{Load convolutional layers from MFS}
        
        \State $\mathcal{V} \gets \text{getVertices}(G_{IR})$ \Comment{Get vertices in $G_{IR}$}
        \State $VT_{dic}=\{\}$ \Comment{dictionary of neighbor node types}
        \For {each vertex $v \in \mathcal{V}$}
            \State $v_{type} \gets \text{Type}(v)$ \Comment{Determine the type of vertex $v$}
            \State $VT_{dic}[v_{type}].append(v)$ \Comment{Append $v$ to the $v_{type}$ list}
        \EndFor
        
        \For {each pair $(V_{Type}, V_{Idx}) \in VT_{dic}$}
        \State $KVS_{VT} \gets  KVS.load(V_{Type})$ \Comment{Load VT chunks}
            \State $nrows,ncols \gets KVS_{VT}.shape()$ \Comment{get rows and cols count}
            \State $VT_{emb} \gets KVS_{VT}.getEmb(V_{Idx})$ \Comment{get embeddings of $V_{Idx}$}
            \State $\widetilde{M}_{VT} \gets \text{SparseTensor}(nrows, ncols, VT_{emb})$ \Comment{Create a sparse tensor of VT embeddings with dimensions nrows and ncols}
        \EndFor        
        \State \Return $\widetilde{M}$ \Comment{Return the constructed model}
    \EndFunction
\end{algorithmic} 
\vspace{-0.5ex}
\end{algorithm}
% \vspace{-5ex}

\subsection{Query-Aware Model Instantiation}
The {\sysName} query-aware inference pipeline, as illustrated in Figure~\ref{figures:KGWISE_InferencePipeline}, begins when a user submits an inference query $\mathcal{Q}_{I}$ containing a set of target nodes $TN$. A corresponding LLM-generated SPARQL query
%\footnote{All the prompts used by {\sysName} are given in the \href{https://github.com/CoDS-GCS/KG-WISE/blob/main/SupplementaryMaterial.pdf}{supplementary material}.}
is retrieved and used to extract the relevant inference subgraph $SG$. In Figure~\ref{figures:KGWISE_InferencePipeline}, gray nodes represent nodes excluded by the query, illustrating how {\sysName} minimizes resource usage by avoiding full-graph loading during inference.
Following the extraction of the inference subgraph, {\sysName} receives the encoded subgraph $SG$ and instantiates a query-specific model $\widetilde{M}$ in two steps:

\nsstitle{- Step 1 (Model Initialization):} The {\sysName} Model Loader initializes the GNN architecture and loads the convolutional weights and biases from the Parameter Store. It builds a mapping from node types and IDs in $SG$ to their corresponding embedding locations in the KV store.\shorten

\nsstitle{- Step 2 (Sparse Tensor Construction):} Instead of loading the full embedding matrix, {\sysName} materializes embeddings only for the nodes that appear in $SG$. These are inserted into a sparse tensor whose indices correspond to the global node encoding space, while all other entries remain empty. This produces a tensor where only $(i,j)$ positions associated with relevant nodes store values, and all non-relevant nodes are implicitly omitted. The resulting sparse tensor becomes the embedding input for $\widetilde{M}$, allowing the forward pass to operate only over nodes in $SG$ rather than the entire KG.

In Algorithm \ref{alg:GNN_Construction}, Lines 2 and 3 initialize the model in memory with the same convolutional layer architecture as that of the model $M$, then load the trained weights and biases. Lines 4 to 8 construct a dictionary where node types are the keys and their corresponding IDs are the values. Lines 9--14 loop through the constructed dictionary, fetching the respective node embeddings from the KV store for each node type. Line 13 generates a sparse tensor with dimensions matching those of the full model $M$, containing only the embeddings of the nodes present in the subgraph. Dense tensors are replaced with sparse tensors to construct the on-demand model $\widetilde{M}$. The complexity of instantiating  $\widetilde{M}$ depends on the number of query nodes $TN$ and the size of $SG$.
% \noindent
\vspace*{-01ex}
\subsection{The Prediction Manager}
% \vspace*{-02ex}
The Prediction Manager is the final module in the inference pipeline, responsible for executing predictions using the  $\widetilde{M}$ and the subgraph $SG$. It dynamically initializes the embeddings for the target nodes ($TN$) while partially loading non-target node embeddings from $\widetilde{M}$. Once initialized, the Prediction Manager processes the subgraph and embeddings through the GNN model's forward pass, which involves aggregation and update operations tailored to the sparse subgraph. Leveraging optimized sparse tensor operations and memory-efficient representations, this process achieves faster computations and up to 60\% reduced memory usage compared to state-of-the-art (SOTA) methods.
Equation \ref{eq_WISE} presents the optimized {\sysName} aggregations for the $TN$ present in the $SG$:
\begin{equation}
\label{eq_WISE}
h_{SG,i}^{(l+1)} = \sigma \left( \sum_{r \in R_{SG}} \sum_{j \in N_{SG,i}^r} \frac{1}{c_{i,r}} W_r^{(l)} h_j^{(l)} + W_0^{(l)} h_i^{(l)} \right)
\end{equation}

In this equation, $h_{SG,i}$ is the feature vector for node $i$ in the query subgraph $SG$, while $R_{SG}$ and $N_{SG}$ represent the subgraph's relations and nodes. Since $|R_{SG}| < |R|$ and $|N_{SG}| \ll |N|$, the message-passing and aggregation steps are smaller and more efficient than in traditional RGCN methods (Equation~\ref{eq_rgcn}).
Consider the set $H$ of node embeddings defined by Equation~\ref{eq_WISE}. As the KG grows, loading $H$ becomes increasingly resource-intensive. To mitigate this, {\sysName} generates $H_{SG}$, a sparse tensor containing embeddings only for the nodes within the inference subgraph $SG$. Since $|H_{SG}| \ll |H|$, this significantly reduces memory consumption and accelerates data loading.
By integrating query-aware inference with a key-value store and RDF engine, {\sysName} constructs an efficient inference pipeline that minimizes memory usage and inference time while preserving model accuracy.

\section{EXPERIMENTAL EVALUATION}
\label{sec:expermients}

This section presents comprehensive experiments to assess our inference system against SOTA GNN inference methods.

\begin{table}[t]
 \vspace*{-2ex}
\addtolength{\tabcolsep}{-0.4em}

\centering
\caption{KGs statistics and GNN tasks: Task Types (TT) include node classification (NC) and link prediction (LP).
}
\vspace*{-2ex}
\label{tbl_exp_tasks}
\small
\begin{tabular}{cccccccc}\hline
TT&Name&KG&\#nodes &\#edges &\#n-type &\#e-type &Metric \\\midrule
NC &PV&MAG-42M&42.4M&166M&58&62&ACC \\
% NC &PD&MAG-42M&Time&87/8/5&Accuracy \\
NC &PC&YAGO4&30.7M&400M&104&98&ACC\\
% NC &CG&YAGO-30M&Random&80/10/10&Accuracy\\
NC &PV&DBLP-15M&15.6M&252M&42&48&ACC\\
% NC &AC&DBLP-15M&Time&80/10/10&Accuracy \\\midrule
LP &AA&DBLP-15M&15.6M&252M&42&48&H@10\\
LP &PO&ogbl-wikikg2&2.5M&17M&9.3K&535&H@10\\
LP &CA&YAGO3-10&123K&1.1M&23&37&H@10\\ \midrule
\end{tabular}
\vspace*{-4ex}
\end{table}

% \begin{table}[t]
% \vspace*{-1ex}
% \centering
% \caption{A summary of our GNN tasks: Task Types (TT) are single-label node classification (NC) or missing entity link prediction (LP).\shorten}
% \vspace*{0ex}
% \label{tbl_exp_tasks}
% \begin{tabular}{cccccc}\hline
% TT&Name&KG&Split&Ratio&Metric \\\midrule
% NC &PV&MAG-42M&Time&84/9/7&Accuracy \\
% % NC &PD&MAG-42M&Time&87/8/5&Accuracy \\
% NC &PC&YAGO-30M&Random&80/10/10&Accuracy\\
% % NC &CG&YAGO-30M&Random&80/10/10&Accuracy\\
% NC &PV&DBLP-15M&Time&79/10/11&Accuracy\\
% % NC &AC&DBLP-15M&Time&80/10/10&Accuracy \\\midrule
% LP &AA&DBLP-15M&Time&99/0.7/0.3&Hits@10\\
% LP &PO&ogbl-wikikg2&Time&94/2.5/3.5&Hits@10\\
% LP &CA&YAGO3-10&Random&99/0.5/0.5&Hits@10\\ \midrule
% \end{tabular}
% \vspace*{-3ex}
% \end{table}

\begin{table*}[t]\centering
\vspace*{-2ex} 
\addtolength{\tabcolsep}{-0.5em}
% \caption{
% Model size and parameter statistics for {\sysName} versus baseline accelerators across various tasks. Our analysis shows that in GNNs trained on KGs, most of model size comes from embeddings of non-target nodes. The baseline accelerator model size remains constant, regardless of the number of target nodes (|$TN$|) in the inference query. {\sysName}'s model size  is dynamic, increasing with larger |$TN$|. ``OOM'' indicates out-of-memory, and ``N/A'' indicates not applicable. Note: For DQ, decoupling weights and embeddings is not possible due to shared quantization parameters and input statistics.}
\caption{
Model size breakdown for {\sysName} and baseline accelerators. In KG-based GNNs, most of the model size comes from non-target node embeddings, which baselines always load in full (constant model size across all $|TN|$). In contrast, {\sysName} loads only embeddings for the extracted subgraph, yielding a query-dependent model size. ``OOM'' denotes out-of-memory and ``N/A'' not applicable. For DQ, weights and embeddings cannot be decoupled due to shared quantization statistics.
}

\vspace*{-1ex}
\label{tbl_model_size}
\small
\begin{tabular}{|c|c|cccc|ccccc|}
\hline
\multirow{2}{*}{\textbf{Task}} &\multirow{2}{*}{\textbf{Model statistics}} &\multicolumn{4}{c|}{\textbf{Baseline Accelerators}} &\multicolumn{5}{c|}{\textbf{{\sysName}'s model with different |$TN$|}}\\ \cline{3-11}
& &\textbf{GS}& \textbf{IBMB}&\textbf{GCNP} &\textbf{DQ} &\textbf{100} &\textbf{200} &\textbf{ 400} &\textbf{800} &\textbf{ 1600} \\ \hline
\multirow{4}{*}{\makecell{DBLP (PV)\\NC}} &\# Parameters &1.86 B&2 B &1.85 B &1.86 B &430.42 M &430.42 M &430.42 M &430.42 M &430.42 M \\
\cline{2-11}
&Size in disk &6.9 GB&7.4 GB&6.9 GB &6.9 GB &17.1 MB &34.8 MB &50.5 MB &67 MB &121.1 MB \\
% \cline{2-11}
% &Size(Weights)/Size(Embeddings)&0.08&0.04&0.05&-&28&12&8&6&3 \\

\cline{2-11}
&Size(Weights)/Size(Model)&0.01&0.01&0.01&-&0.5&0.33&0.2&0.11&0.06 \\

\cline{2-11}
&Size(Embeddings)/Size(Model)&0.99&0.99&0.99&-&0.5&0.66&0.79&0.88&0.93 \\
\Xhline{1.5\arrayrulewidth}

\multirow{4}{*}{\makecell{YAGO4 (PC)\\ NC}} &\# Parameters &3.65 B &B&1.83 B &3.65 B &243.7 M &243.7 M &243.7 M &243.7 M &243.7 M \\
\cline{2-11}
&Size in disk &13.6 GB &13.9 GB&13.6 GB &13.6 GB &18.4 MB &24.1 MB &31 MB &46.4 MB &81.5 MB \\
\cline{2-11}
% &Size(Weights)/Size(Embeddings)&0.12&0.08&0.06&-&371&151&89&45&22\\
% \cline{2-11}
&Size(Weights)/Size(Model)&0.01&0.01&0.01&-&0.95&0.86&0.68&0.45&0.35 \\

\cline{2-11}
&Size(Embeddings)/Size(Model)&0.99&0.99&0.99&-&0.04&0.13&0.31&0.54&0.64 \\

\Xhline{1.5\arrayrulewidth}
\multirow{5}{*}{\makecell{MAG (PV)\\ NC}} &\# Parameters &5.35 B &B&OOM &5.35 B &876.26 M &876.26 M &876.26 M &876.26 M &876.26 M \\
\cline{2-11}
&Size in disk &19.9 GB &20.3 GB &OOM &19.9 GB &17.1 MB &34.8 MB &50.5 MB &67 MB &121.1 MB \\
\cline{2-11}
% &Size(Weights)/Size(Embeddings)&0.12&0.04&OOM&-&194&48&29&20&10 \\
% \cline{2-11}
&Size(Weights)/Size(Model)&0.01&0.01&0.01&-&0.19&0.17&0.14&0.09&0.05 \\

\cline{2-11}
&Size(Embeddings)/Size(Model)&0.99&0.99&0.99&-&0.80&0.82&0.85&0.90&0.94 \\

% \hline
\Xhline{3\arrayrulewidth}
\multirow{4}{*}{\makecell{DBLP (AA)\\ LP}} &\# Parameters &N/A&N/A&N/A &OOM &49.79 M &49.79 M &49.79 M &49.79 M &49.79 M \\
\cline{2-11}
&Size in disk &N/A&N/A &N/A &OOM &1.9 MB &3.7 MB &7.1 MB &14.1 MB &27.3 MB \\
\cline{2-11}
% &Size(Weights)/Size(Embeddings)& - &-& - & - &43&18&7&4&2 \\
% \cline{2-11}
&Size(Weights)/Size(Model)&-&-&-&-&0.3&0.15&0.07&0.03&0.02 \\
\cline{2-11}
&Size(Embeddings)/Size(Model)&-&-&-&-&0.69&0.84&0.92&0.96&0.97 \\

\Xhline{1.5\arrayrulewidth}
\multirow{4}{*}{\makecell{YAGO3 (CA)\\ LP}} &\# Parameters &N/A&N/A &N/A &1.50 M &294.53 K &294.53 K &294.53 K &294.53 K &294.53 K \\
\cline{2-11}
&Size in disk &N/A &N/A&N/A &69 MB &3 MB &3.6 MB &4.2 MB &4.7 MB &4.8 MB \\
\cline{2-11}
% &Size(Weights)/Size(Embeddings)&-& - & - & - &7&6&5&4&4\\
% \cline{2-11}
&Size(Weights)/Size(Model)&-&-&-&-&0.12&0.10&0.09&0.08&0.08 \\

\cline{2-11}
&Size(Embeddings)/Size(Model)&-&-&-&-&0.87&0.89&0.90&0.91&0.92 \\
\Xhline{1.5\arrayrulewidth}
\multirow{4}{*}{\makecell{WikiKG (PO)\\LP}} &\# Parameters &N/A&N/A &N/A &OOM &49.79 M &49.79 M &49.79 M &49.79 M &N/A \\
\cline{2-11}
&Size in disk &N/A &N/A&N/A &OOM &8.8 MB &10.4 MB &12.6 MB &16.6 MB & - \\
% \cline{2-11}
% &Size(Weights)/Size(Embeddings)& -&- & - & - &388&200&125&72& - \\
\cline{2-11}
&Size(Weights)/Size(Model)&-& - & - & - &0.21&0.12&0.08&0.04& - \\
\cline{2-11}
&Size(Embeddings)/Size(Model)&-& - & - & - &0.78&0.87&0.91&0.95& - \\
\hline
\end{tabular}
\vspace*{-2ex}
\end{table*}

\begin{figure*}[t]
\vspace*{-2ex}
     \centering
     % \hfill
     \begin{subfigure}
         \centering
         \includegraphics[width=0.99\textwidth]
         {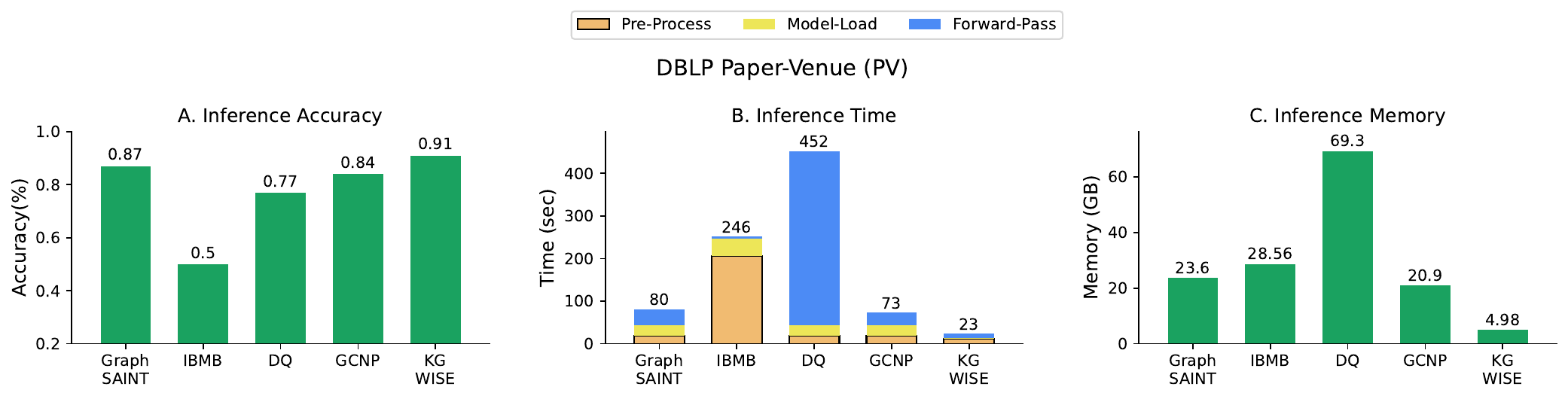}
         %{figures/Experiments/NC_MAG/MAG_INF_combined_plots.pdf}
         % \caption{DBLP}
         \label{fig:DBLP-15M-RW}
         %\label{fig:MAG-42M-RW}
         \vspace*{-1ex}
         
     \end{subfigure}
     % \hfill
     \begin{subfigure}
         \centering 
         \includegraphics[width=0.99\textwidth]
         {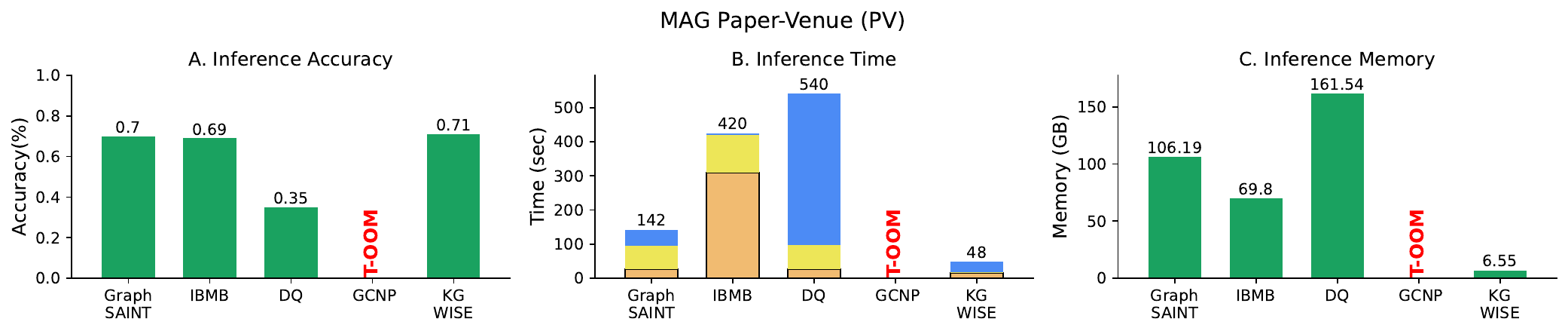}
         %{figures/Experiments/NC_DBLP/DBLP_INF_combined_plots.pdf}        
         % \caption{MAG}
         \label{fig:MAG-42M-RW}
         %\label{fig:DBLP-15M-RW}
         \vspace*{-1ex}
     \end{subfigure}
     \begin{subfigure}
         \centering         \includegraphics[width=0.99\textwidth]{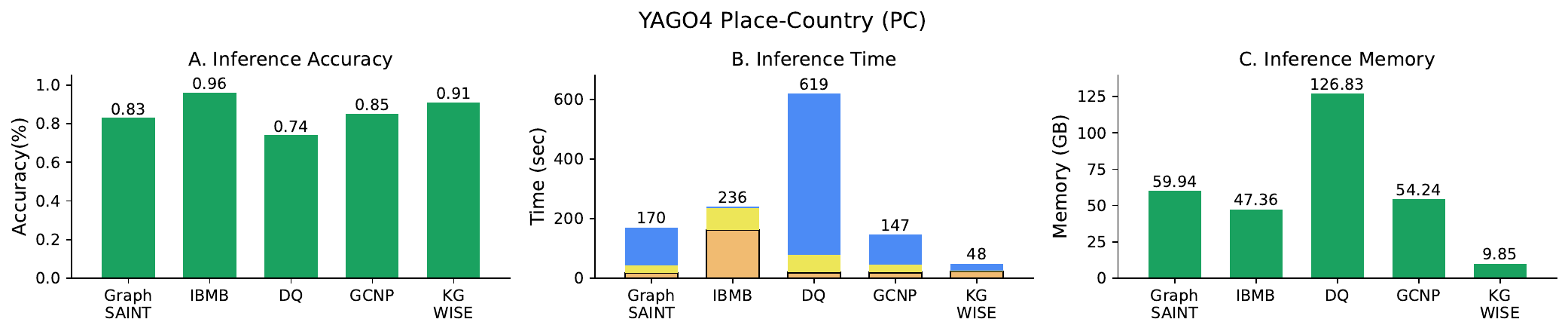}
         % \caption{YAGO-30M (Place-Country)}
         \label{fig:YAGO-30M-RW}
     \end{subfigure}
     \vspace*{-5ex}
    \caption{ 
    Performance across NC tasks is based on three metrics: (A) Inference Accuracy (higher is better), (B) Inference-Time (lower is better), and (C) Inference Memory (lower is better). The top and middle sections illustrate the results for the Paper-Venue task on DBLP and MAG, respectively. The bottom figures present the results of the Place-Country task on YAGO4. The inference query performs inference for 1K target nodes stratified across all classes.  {\sysName} archives comparable inference accuracy compared with the SOTA training/inference accelerators (Graph SAINT, IBMB, GCNP, and DQ). {\sysName} outperforms the SOTA methods by up to 28x in inference time on YAGO4 with memory reduction up to 98\%. \shorten
    } 
\label{fig:KGWISE_NC}
  \vspace*{-3ex}
\end{figure*}

\subsection{Evaluation Setup}
\noindent\textbf{Benchmark Datasets and GNN Tasks:}
We utilize the KGBen benchmark~\cite{KGTOSA}, a large and challenging benchmark with KGs of up to 42 million nodes and 166 million edges. We also used small KGs as some methods do not scale to the large ones. 
We employ several node classification (NC) and link prediction (LP) tasks, each containing from 100 to 1600 target nodes, to simulate real large-scale workloads.
%We also used several GNN node classification and link prediction tasks. 
In total, we use six real-world KGs from diverse application domains, such as DBLP~\cite{KG_DBLP}, MAG~\cite{MAG}, and YAGO4~\cite{KG_Yago4}.
%, and ogbl-wikikg2~\cite{KG_Wikidata}.
% of the used KGs and clarify the used GNN tasks.  
Given the high computational demands of existing GNN methods for link prediction tasks on large KGs, we incorporate smaller datasets like YAGO3-10 ~\cite{Yago3-10}, a more compact version of YAGO3, and ogbl-wikikg2 ~\cite{KG_Wikidata}, a dataset derived from Wikidata. 
%Table~\ref{tbl_exp_tasks} summarizes the statstics.
%This smaller dataset includes 2.5 million entities and 535 edge types. 

Our evaluation encompasses both NC and LP tasks. For NC tasks, we focus on single-label classifications, following established heterogeneous graph datasets from previous studies~\cite{OGB,SeHGNN,Shadow-GNN}. We use accuracy as the primary metric to evaluate performance on these NC tasks. We choose the Hits@10 metric to evaluate LP
%the predictions ranking
performance following SOTA methods~\cite{MorsE,OGB,Shadow-GNN}. 
The train–validation–test splits are derived either from three KG versions across timesteps or from random splits. For each dataset, we follow the KGBen benchmark settings~\cite{KGTOSA}.
Table~\ref{tbl_exp_tasks} provides statistics on the KG datasets , the node classification (NC) and link prediction (LP) tasks we used.

\noindent\textbf{The GNN Baseline Inference Methods:} 
We evaluate {\sysName} on node classification (NC) tasks using RGCN-based GNNs with training accelerators like GraphSAINT~\cite{GraphSAINT} and IBMB~\cite{ibmb}, and compare it against adapted inference accelerators including Graph Channel Pruning (GCNP)~\cite{GCNP}, Degree-Quant (DQ)~\cite{Degree-Quant}, and GKD~\cite{GKD}. Since the original implementations of GCNP, DQ, and GKD support only homogeneous graphs, we extended them to work with RGCN for heterogeneous KGs. Specifically, we replaced linear layers with quantized versions in DQ, applied lasso-based pruning from GCNP, and used GKD’s distillation loss in RGCN.
%
% Figure~\ref{fig:KGWISE_NC} shows our adaptation of GCNP significantly reduces inference time and memory compared to the unpruned GraphSAINT model. 
For link prediction (LP), we compare {\sysName} against DQ (as an inference accelerator) and MorsE~\cite{MorsE} (as a training accelerator), both adapted to support RGCN. We use the default tuned training parameters provided by each method's original implementation.
Furthermore, we compare {\sysName} with {\TOSA}~\cite{KGTOSA} to analyze the effects of subgraph extraction and inference optimization in Section~\ref{sec:ablation}.

\noindent\textbf{Computing Infrastructure:} 
All inference experiments were executed on an Ubuntu VM with dual 64-core Intel Xeon 2.4\,GHz CPUs and 256\,GB RAM. Virtuoso 07.20.3229 served as the RDF engine, with one instance per KG. A separate VM was used per task for training and inference, and node/edge embeddings were stored in Zarr-Python~\cite{ZarrPython} (v2.17.1). We evaluated {\sysName} using both proprietary LLMs (Gemini 2.5 Flash, ChatGPT~4o, ChatGPT~5) accessed via public APIs, and open-weight LLMs (GPT-oss-20B, DeepSeek-R1-Distill-Qwen-14B, Qwen3-30B-A3B-Instruct-2507) hosted locally on a Compute Canada 16\,GB VRAM vGPU via llama-cpp. A 32\,GB Nvidia Volta V100 GPU was used for GPU-based inference evaluation.

\noindent\textbf{{\sysName} Implementation and Settings:}
{\sysName}\footnote{{\sysName} is open-sourced at \url{https://github.com/CoDS-GCS/KG-WISE}} is implemented in Python 3.8, using PyTorch~2.1 and PyG~2.5. Model parameters are persisted in a file store, while metadata is maintained in an RDF graph. Node embeddings are Xavier-initialized. GraphSAINT~\cite{GraphSAINT} is used for NC tasks and RGCN~\cite{RGCN} for LP tasks. The LLM-guided sampler supports both proprietary and open-weight LLMs; Gemini 2.5 Flash was used for the reported experiments. The hop limit $K$ is fixed to 2 across tasks to avoid over-smoothing~\cite{GNN_Over_Smoothing}. 
SPARQL queries are executed in parallel: the target nodes of each inference query are partitioned into batches and retrieved using multi-threaded SPARQL execution.

\noindent\textbf{Why Full-Model Loading Is Wasteful:}
Table~\ref{tbl_model_size} shows that, for all baseline accelerators, embeddings consistently account for $\sim99\%$ of the total model size, while weights contribute only $\sim1\%$, regardless of $|TN|$.
%As a result, 
This is significant since baselines always load the full model (e.g., 6.9--19.9\,GB in DBLP/MAG/YAGO4), even when the query touches only a tiny fraction of the graph. In contrast, {\sysName} materializes only the subgraph-specific embeddings: for DBLP NC the model shrinks from 6.9\,GB to as little as 17\,MB (at $|TN|{=}100$), and even at $|TN|{=}1600$ remains just 121\,MB. Similar reductions appear in MAG (19.9\,GB$\rightarrow$17--121\,MB) and YAGO4 (13.6\,GB$\rightarrow$18--81\,MB). These observations confirm that most of the footprint in KG-GNNs is query-irrelevant, and that partial loading enables two orders of magnitude smaller model instantiation, which we next quantify in memory and latency.

\subsection{Node Classification}
This experiment evaluates the performance of SOTA methods under inference workloads for node classification, as shown in Figure~\ref{fig:KGWISE_NC}. We include two SOTA GNN training methods, GraphSAINT~\cite{GraphSAINT} and IBMB~\cite{ibmb}, and two inference accelerators, GCNP~\cite{GCNP} and DQ~\cite{Degree-Quant}. We also attempted to run the GKD~\cite{GKD} inference accelerator, but it could not scale to our large graphs and ran out of memory in all our experiments. 
The inference query targets 1K nodes, distributed across all task classes.
 
\begin{figure}[t]
\vspace*{-2ex}
     \centering
     % \hfill
     \begin{subfigure}
         \centering 
         \includegraphics[width=0.5\textwidth]
         {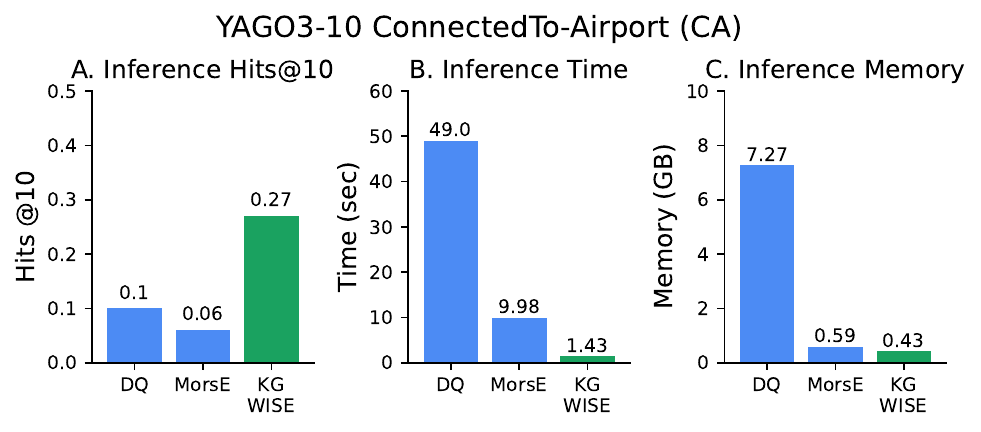}
         \label{fig:LP_YAGO}
         \vspace*{-3ex}
     \end{subfigure}
     \hfill     
      \begin{subfigure}
         \centering
         \includegraphics[width=0.5\textwidth]
         {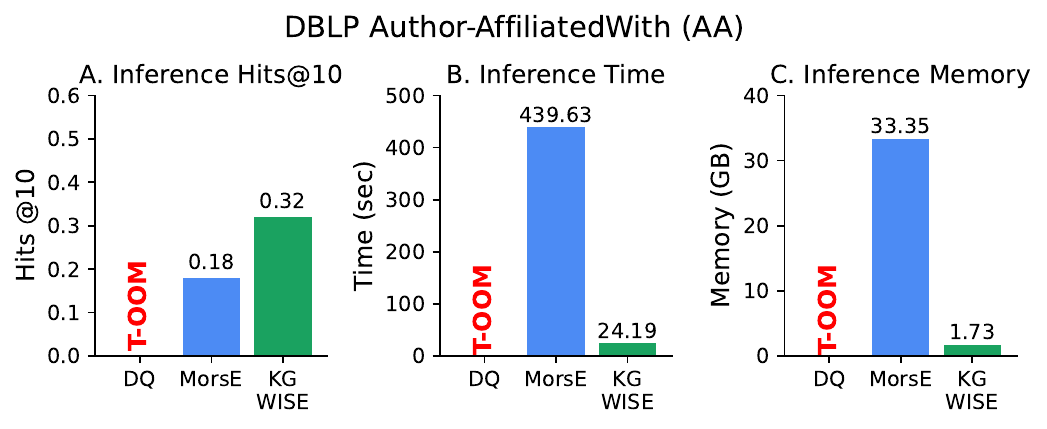}
         % \caption{DBLP}
         \label{fig:LP_DBLP}
         %\label{fig:MAG-42M-RW}
         \vspace*{-3ex}         
     \end{subfigure}
     \begin{subfigure}
         \centering
         \includegraphics[width=0.5\textwidth]         {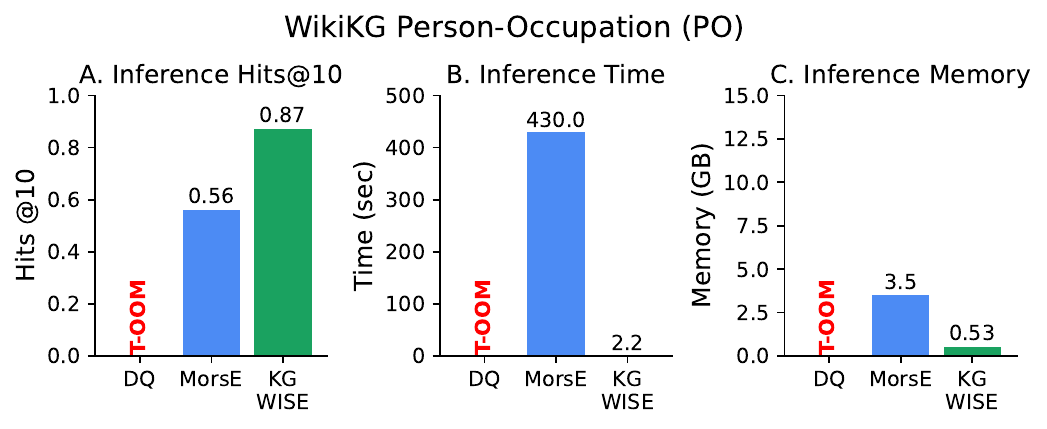}
         % \caption{YAGO-30M (Place-Country)}
         \label{fig:LP_WIKIKG}
     \end{subfigure}
     \vspace*{-5ex}
    % \caption{ 
    %    Performance across LP tasks is based on three metrics: (A) Inference Hits@10 (higher is better), (B) Inference time (lower is better), and (C) Inference Memory (lower is better). 
    %    Top row: YAGO3 Airport-ConnectedTo task; middle: DBLP Author-Affiliation; bottom: Wiki-KG Person-Occupation. Each query performs inference on 100 target nodes across source types. {\sysName} achieves higher accuracy and up to 7× faster inference than MorsE on YAGO3-10.
    %    } 
    \caption{
Results on link prediction tasks across three KGs, reporting (A) Hits@10, (B) inference time, and (C) memory usage. Rows correspond to YAGO3-10 (CA), DBLP (AA), and WikiKG (PO). Each query uses 100 target nodes. {\sysName} improves accuracy while reducing inference time (up to 7$\times$) and memory compared to MorsE and DQ (which fails OOM in two cases).
}

\label{fig:KGWISE_LP}
  \vspace*{-3ex}
\end{figure}

{\sysName} achieves comparable or up to 4\% higher accuracy across three tasks. This is attributed to our LLM-guided task-oriented subgraph, which yields a smaller and sparser model. The subgraph's sparsity acts as a form of regularization, improving accuracy by excluding irrelevant nodes and edges.
{\sysName} achieved up to a 22× speedup in inference time and a 93\% reduction in memory usage on the DBLP-PV task. On the MAG-PV task, the largest and densest graph, {\sysName} delivered a 10× faster inference and 92\% memory savings compared to DQ, while GCNP faced OOM errors. For the YAGO4-PC task, which involves complex node and edge types for place-country classification, {\sysName} achieved a 28× speedup with 98\% memory savings over DQ, and a 6× speedup with 93\% savings over GCNP.

Training accelerators like GraphSAINT and IBMB achieved the highest accuracy across all tasks but incurred higher sampling overhead and require full model loading into memory. In comparison, {\sysName} maintained competitive accuracy while 
achieving faster loading, quicker inference, and lower memory footprint.
This efficiency stems from the query-specific model constructed during inference, which is substantially smaller, as shown in Table~\ref{tbl_model_size}.
Overall, tasks with higher graph sparsity benefited most, aligning with typical characteristics of real-world knowledge graphs~\cite{KG-Sparsity}. 

\subsection{Link Prediction}
This experiment compares the performance of {\sysName} to two SOTA methods, DQ and MorsE~\cite{MorsE}, on inference workloads for link prediction on KGs from diverse domains. The results are shown in Figure \ref{fig:KGWISE_LP}. The inference query consists of 100 target nodes, distributed across different source node types. {\sysName} achieved higher accuracy than 
DQ and MorsE in all tasks while being
significantly faster and more resource-efficient.
On the DBLP dataset, the largest of the three KGs in this experiment, {\sysName} outperforms MorsE by achieving 17 points higher Hits@10, 18× faster inference, and 95\% lower memory usage. 
Since link prediction assigns a score to every node in the subgraph, larger graphs increase both noise and computational cost. {\sysName}'s  subgraph sampler improves accuracy while simultaneously reducing overhead by extracting semantically relevant subgraphs for each query node, thereby focusing the computation on only relevant nodes.
A similar pattern is observed in the YAGO3-10-CA dataset, where {\sysName} delivers a 7× speedup and 28\% memory reduction over MorsE.

% WikiKG presents the most heterogeneous KG structure, with the highest number of node and edge types, and hence the highest number of semantically irrelevant entities that can degrade performance. {\sysName}'s LLM-based sampler effectively prunes such noise, reducing relation types from 535 to just 50. For example, when predicting a person's occupation, attributes related to digital games, such as input devices or programming languages, are excluded. In contrast, the entity representing the digital game itself is retained to capture the person's interests, which is helpful in predicting occupation. This targeted sampling boosts inference, yielding a 31-point increase in Hits@10 over MorsE, a 195× speedup, and 85\% memory savings.

WikiKG is the most heterogeneous KG in our benchmark, with many node and edge types that introduce semantically irrelevant neighbors. {\sysName}'s LLM-guided sampler prunes this noise, reducing 535 relation types to just 50. For the Person–Occupation task, it excludes distractors such as input-device or programming-language relations while retaining only informative ones (e.g., digital game entities that reflect user interests). This targeted reduction yields a 31-point increase in Hits@10, a 195$\times$ speedup, and an 85\% memory savings over MorsE.\shorten

\subsection{Ablation Study}
\label{sec:ablation}

\begin{figure}[t]
\vspace*{-2ex}
     \centering
     % \hfill
     \begin{subfigure}
         \centering
         \includegraphics[width=0.48\textwidth]
         {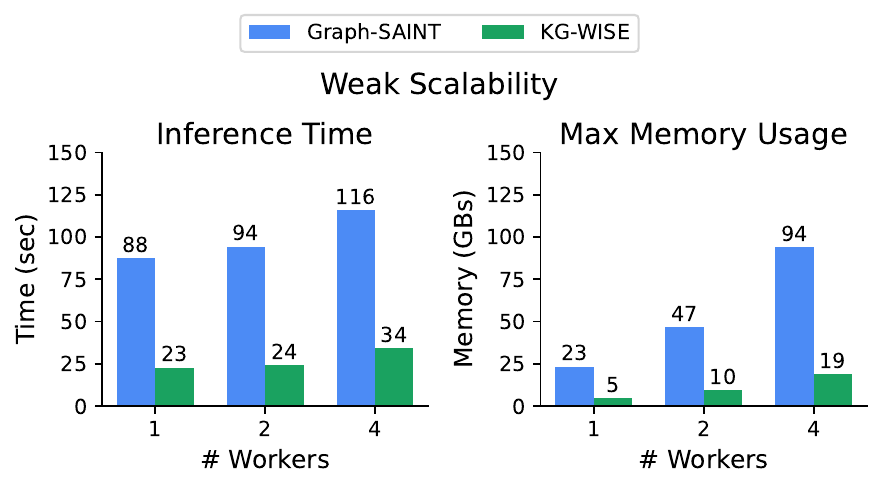}
         \label{fig:weak_sca}
         \vspace*{-2ex}         
     \end{subfigure}
     % \hfill
     \begin{subfigure}
         \centering 
         \includegraphics[width=0.48\textwidth]{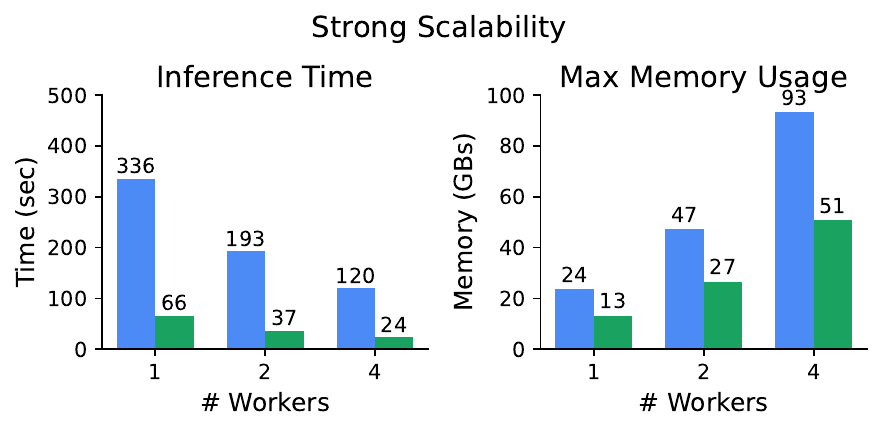}
         \label{fig:strong_sca}
         %\label{fig:DBLP-15M-RW}
         \vspace*{-2ex}
     \end{subfigure}
     \vspace*{-4ex}
    \caption{The inference time and memory consumption of {\sysName} and Graph-SAINT inference pipelines under weak and strong scalability settings. {\sysName} shows better scalability in both settings.} 
    \label{fig:KGWISE_Scalability}
    \vspace*{-2ex}
\end{figure}

\noindent\textbf{{\sysName} Scalability:}
We demonstrate the scalability of {\sysName} through weak and strong scalability tests as shown in Figure \ref{fig:KGWISE_Scalability}.
In the weak scalability setting, using the DBLP inference dataset, both the problem size (number of queries) and the number of workers (CPU cores) increase proportionally. Each worker operates independently with its own model copy and processes the entire inference pipeline. {\sysName} achieves a time efficiency of 0.68 compared to 0.76 for Graph-SAINT, with better memory efficiency at 0.26 versus 0.24.\\ In the strong scalability setting, a constant problem size consisting of inference queries from NC tasks namely [DBLP (\#200), YAGO (\#200), MAG (\#200),  DBLP (\#400), YAGO (\#400), MAG (\#400),  DBLP (\#800), YAGO (\#800), MAG (\#800)] is tested while varying the number of workers (1 to 4, with 4 cores per worker). At 4 workers, {\sysName} achieves a speedup of 2.4 compared to 1.83 for Graph-SAINT, with memory efficiency of 0.25 compared to 0.49. In summary, {\sysName} demonstrates better memory savings in memory-bound settings (weak scalability) and higher computational speed in CPU-bound settings (strong scalability).

\begin{figure}[htbp]
    \centering
    \begin{minipage}{0.45\textwidth}
        \centering
     \begin{subfigure}
         \centering
         \includegraphics[width=0.99\textwidth]
         {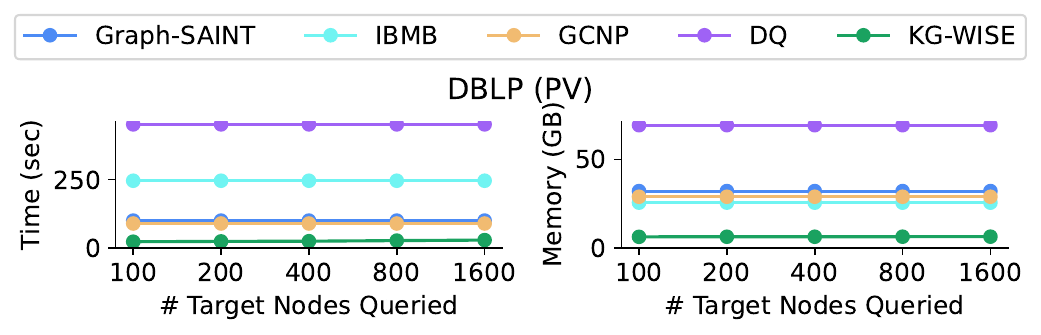}
         % \caption{DBLP}
         \label{fig:NC_var_DBLP}
         %\label{fig:MAG-42M-RW}
         \vspace*{-2ex}
         
     \end{subfigure}
     % \hfill
     \begin{subfigure}
         \centering 
         \includegraphics[width=0.99\textwidth]         {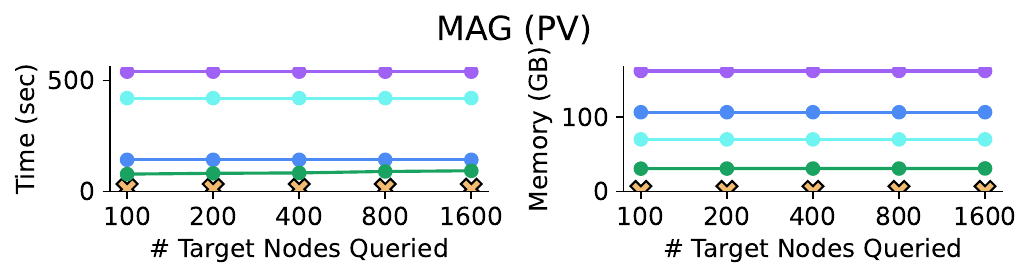}
         \label{fig:NC_var_MAG}
         %\label{fig:DBLP-15M-RW}
         \vspace*{-2ex}
     \end{subfigure}
     \begin{subfigure}
         \centering
         \includegraphics[width=0.99\textwidth]         {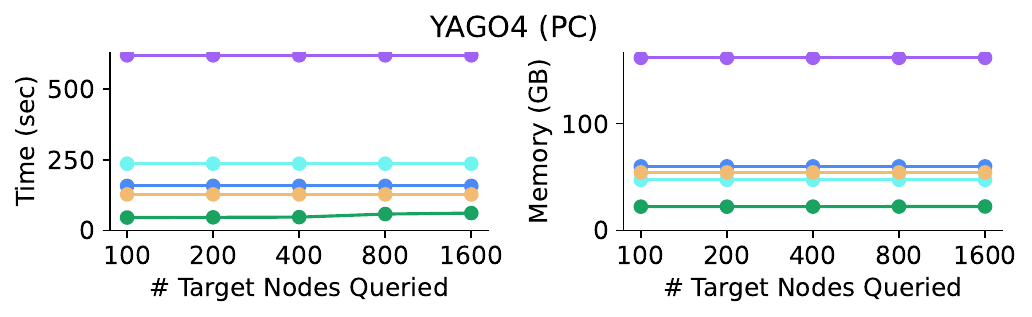}
         % \caption{YAGO-30M (Place-Country)}
     \end{subfigure}
    \caption{
    % The Inference time and memory of {\sysName} Versus SOTA GNN accelerators while varying the number of target nodes per query. The top row represents the DBLP NC task, the middle row represents the MAG NC task, and the bottom row represents the YAGO NC task. The existing method's inference time and memory are almost constant regardless of the inference query size. {\sysName} inference time and memory consumption are sub-linear to the number of target nodes per query which makes {\sysName} a query-aware on-demand service. 
    Node classification inference time and memory usage of {\sysName} compared to SOTA GNN accelerators as the number of target nodes per query increases. The top, middle, and bottom rows correspond to the DBLP, MAG, and YAGO NC tasks, respectively. {\sysName} exhibits superior performance in all cases.}
    % While baseline methods show nearly constant resource usage, {\sysName} scales sub-linearly with query size due to our efficient query-aware inference.
    % } 
    \label{fig:KGWISE_ByTargetNodes}
\end{minipage}
 \hspace{0.02\textwidth}
 \begin{minipage}{0.45\textwidth}
     \centering
     % \hfill
          \begin{subfigure}
         \centering
         \includegraphics[width=0.99\textwidth]       {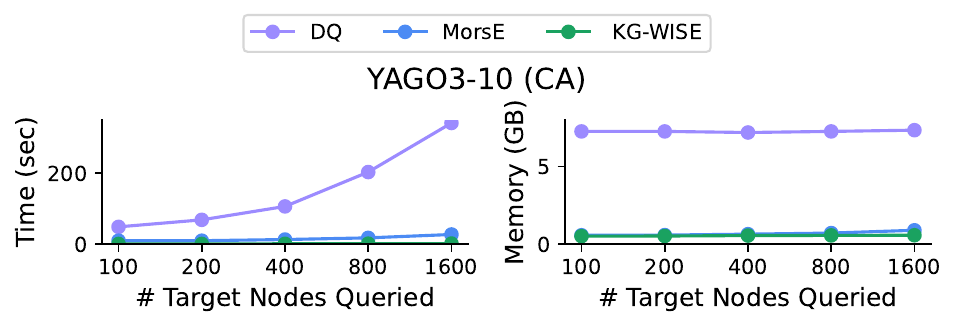}
         % \caption{YAGO-30M (Place-Country)}
         \vspace*{-2ex}
     \end{subfigure}
     \begin{subfigure}
         \centering
         \includegraphics[width=0.99\textwidth]
         {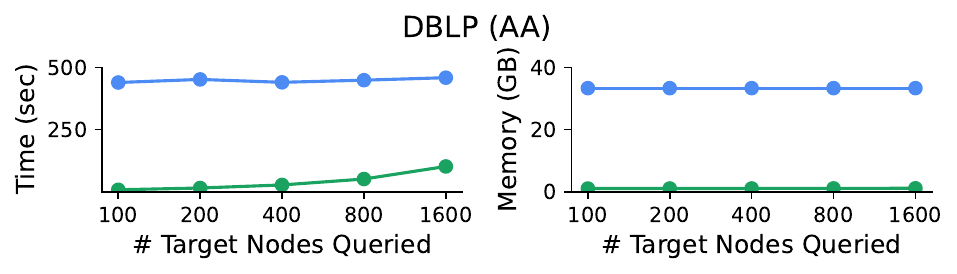}
         % \caption{DBLP}
         \label{fig:NC_var_DBLP}
         %\label{fig:MAG-42M-RW}
            \vspace*{-2ex}
     \end{subfigure}
     % \hfill
     \begin{subfigure}
         \centering          \includegraphics[width=0.99\textwidth]{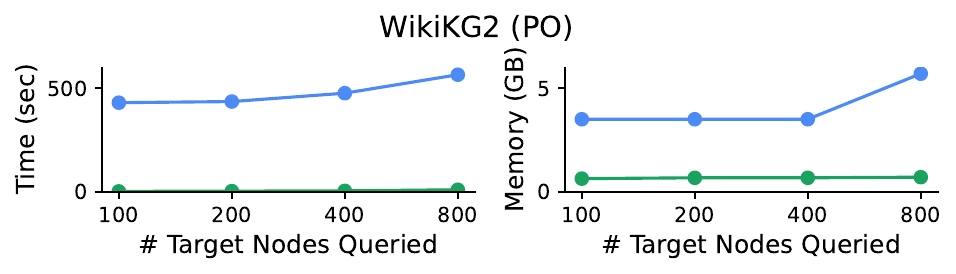}
         \label{fig:NC_var_MAG}
         %\label{fig:DBLP-15M-RW}
         % \vspace*{-2.3ex}
     \end{subfigure}
    \caption{Link prediction inference time and memory usage of {\sysName} vs. SOTA GNN accelerators across varying numbers of target nodes per query. The top, middle, and bottom rows correspond to the YAGO3-10-CA, DBLP-PV, and WikiKG-PO LP tasks, respectively. Baseline methods show near-linear inference time and constant memory usage, regardless of query size. In contrast, {\sysName} exhibits sub-linear growth in both time and memory, enabling efficient, query-aware, on-demand inference.}
    \label{fig:KGWISE_ByTargetNodes_LP}

\end{minipage}
\end{figure}

\noindent\textbf{{\sysName} under varying inference-query size.}
{\sysName} is most effective when the inference query contains a limited number of target nodes. Figures \ref{fig:KGWISE_ByTargetNodes} and \ref{fig:KGWISE_ByTargetNodes_LP} in the 
% \href{https://github.com/CoDS-GCS/KG-WISE/blob/main/SupplementaryMaterial.pdf}
\ref{sec:appendix} results for node classification and link prediction as $|TN|$ increases. For baselines, both inference time and memory remain almost constant because they always load the full model and graph. In contrast, {\sysName} grows only slightly with $|TN|$, since it loads only embeddings for the extracted subgraph. For example, on DBLP, inference time rises from 25\,s to 30\,s when $|TN|$ increases from 100 to 1.6K, while the fastest baseline (GCNP) stays around 150\,s. Memory usage shows a similar trend: at 1.6K target nodes, {\sysName} uses 6.3\,GB versus 30\,GB for GCNP (79\% reduction). On denser KGs such as MAG and YAGO, the ratio of embedding size to weights is lower (Table~\ref{tbl_model_size}), which narrows the margin, but {\sysName} still outperforms all baselines in both time and memory.
For link prediction, the inference time and memory consumption of other methods are constant or increase linearly with the number of target nodes.
In contrast, {\sysName} shows sub-linear scaling in both inference time and memory usage.

% \noindent\textbf{{\sysName} on GPU}
% We evaluate the performance of {\sysName} and baseline during inference on a CPU versus a GPU. {\sysName} effectively utilizes GPU hardware to accelerate inference. For instance, in the DBLP dataset, the total inference time decreased from 21 seconds on CPU to 13 seconds on GPU. Whereas the GSaint even on GPU is slower than {\sysName} on CPU. Similarly, for the MAG dataset, the time was reduced significantly from 62 seconds to just 22 seconds. In the case of YAGO4, inference time dropped from 34 seconds to 15 seconds.
% While the GPU memory consumption of {\sysName} is higher than that of the CPU across all datasets, this is expected due to the inherently different memory management architectures between system RAM and GPU memory. These two types of memory are not directly comparable in terms of utilization. Nevertheless, we report the GPU memory usage to provide a complete picture of the {\sysName} footprint on GPU hardware. 

\begin{figure}[t]
\vspace*{-2ex}
     \centering
     % \hfill
     \begin{subfigure}
         \centering
         \includegraphics[width=0.48\textwidth]
         {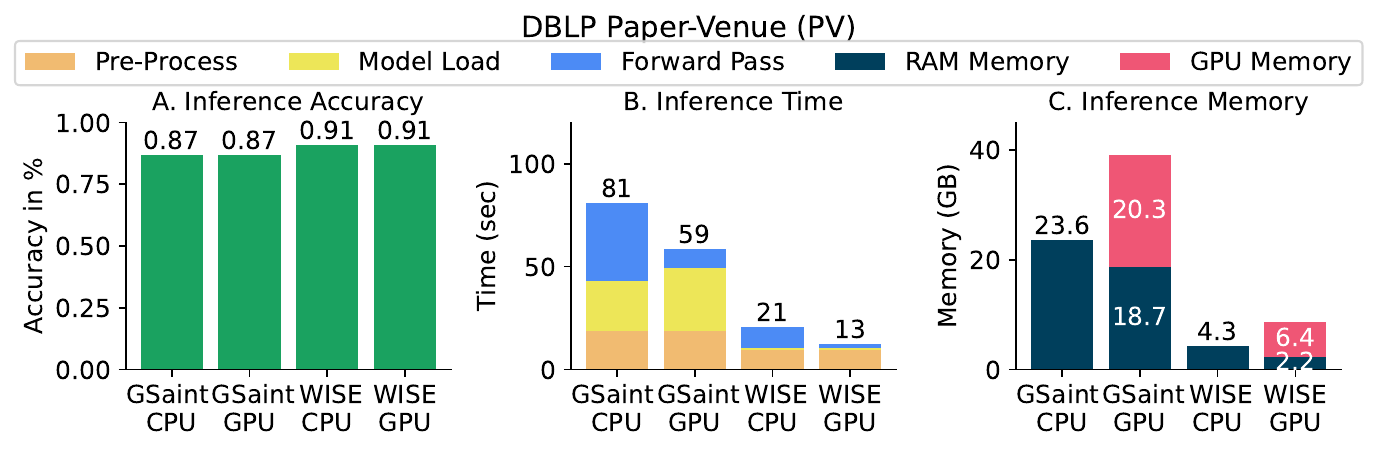}
         \label{fig:WISE_GPU_DBLP}
         % \vspace*{-2ex}         
     \end{subfigure}
     % \hfill
     \begin{subfigure}
         \centering 
         \includegraphics[width=0.48\textwidth]{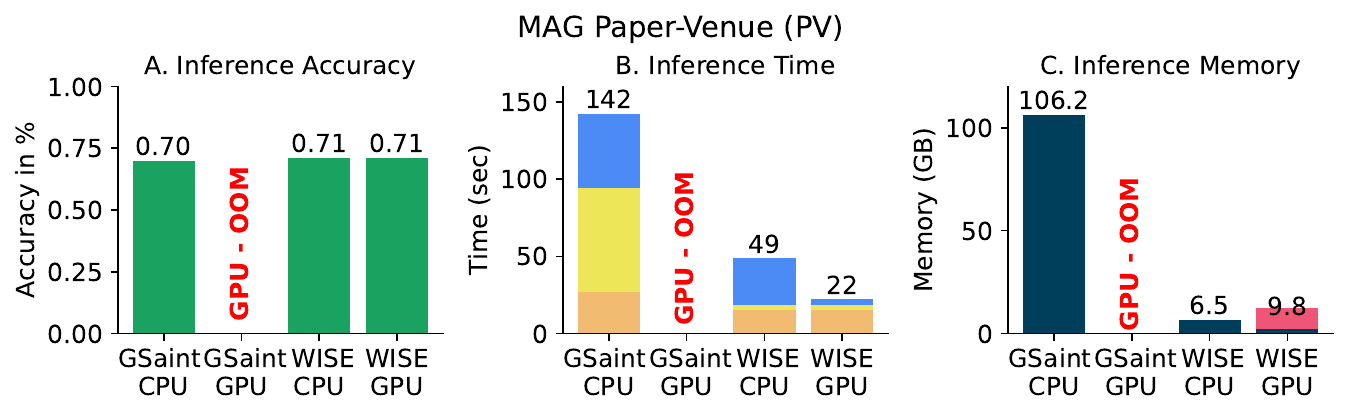}
         \label{fig:WISE_GPU_MAG}
         % \vspace*{-2ex}         
     \end{subfigure}
     \begin{subfigure}
         \centering
         \includegraphics[width=0.48\textwidth]       {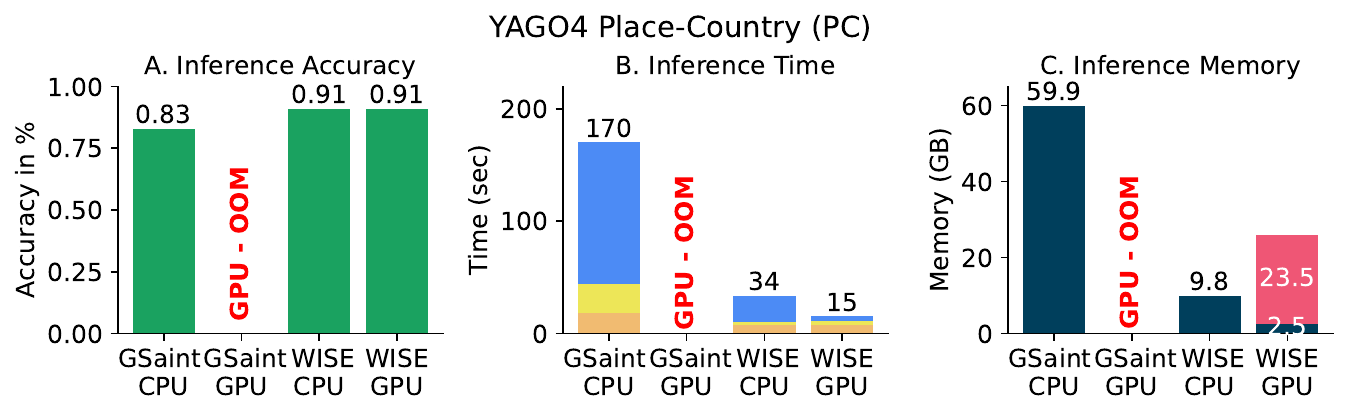}
         \label{fig:WISE_GPU_YAGO}
     \end{subfigure}
     \vspace*{-4ex}
    \caption{ Evaluation of {\sysName} while running on a CPU and a GPU} 
    \label{fig:WISE_GPU}
    \vspace*{-3ex}
\end{figure}

\noindent\textbf{{\sysName} on GPU.}
We evaluate the performance of {\sysName} and baseline during inference on a CPU versus a GPU. {\sysName} effectively utilizes GPU hardware to accelerate inference. Figure~\ref{fig:WISE_GPU} compares CPU and GPU inference. {\sysName} benefits directly from GPU acceleration: on DBLP, inference time drops from 21\,s (CPU) to 13\,s (GPU); on MAG from 49\,s to 22\,s; and on YAGO4 from 34\,s to 15\,s. In contrast, GraphSAINT (GSAINT) remains slower even when moved to GPU, and in several LP/NC tasks the baseline models run out of GPU memory. {\sysName} avoids these failures by loading only subgraph-specific embeddings, which keeps its GPU footprint small enough to fit within VRAM. GPU memory usage is higher than CPU RAM usage, as expected, but remains far below the baselines and does not limit execution. These results show that {\sysName} can leverage GPU hardware effectively while retaining its partial-loading advantage.

\begin{figure}[t]
\vspace*{-2ex}
     \centering
     % \hfill
     \begin{subfigure}
         \centering
         \includegraphics[width=0.48\textwidth]
         {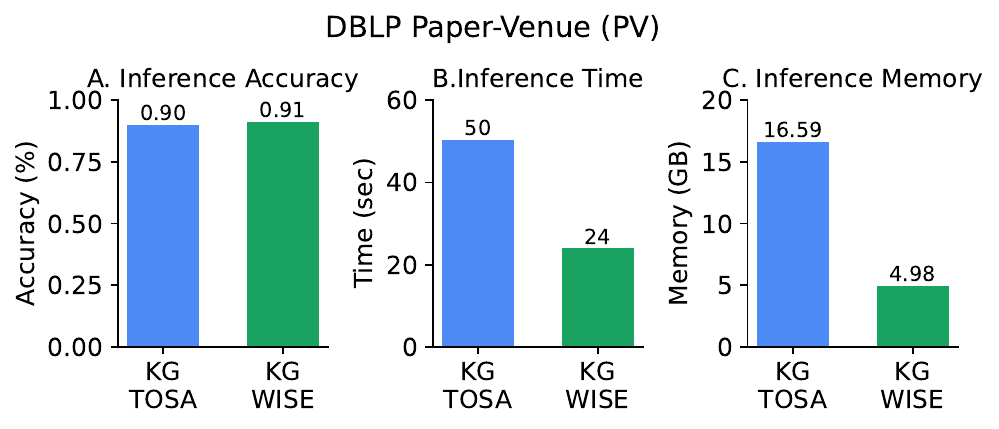}
         % \caption{DBLP}
         \label{fig:NC_KG-TOSA_WISE_DBLP}
         %\label{fig:MAG-42M-RW}
         % \vspace*{-3ex}         
     \end{subfigure}
     % \hfill
     \begin{subfigure}
         \centering 
         \includegraphics[width=0.48\textwidth]{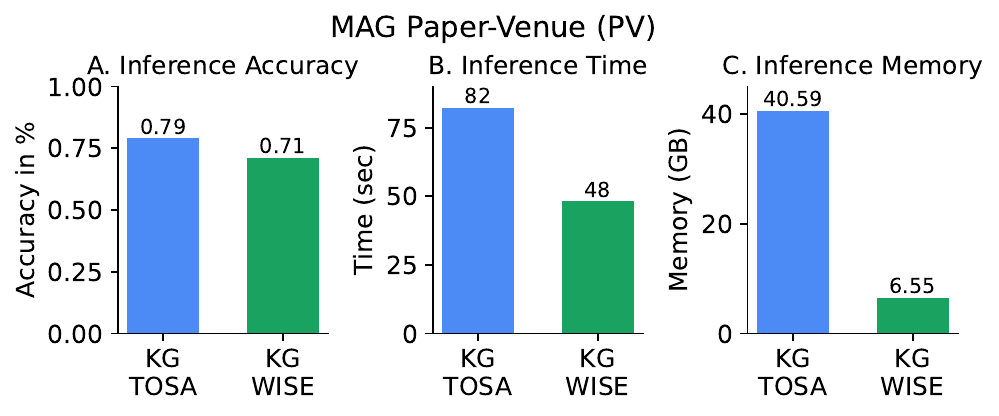}
         \label{fig:NC_KG-TOSA_WISE_MAG}
         %\label{fig:DBLP-15M-RW}
         % \vspace*{-3ex}         
     \end{subfigure}
     \begin{subfigure}
         \centering
         \includegraphics[width=0.48\textwidth]       {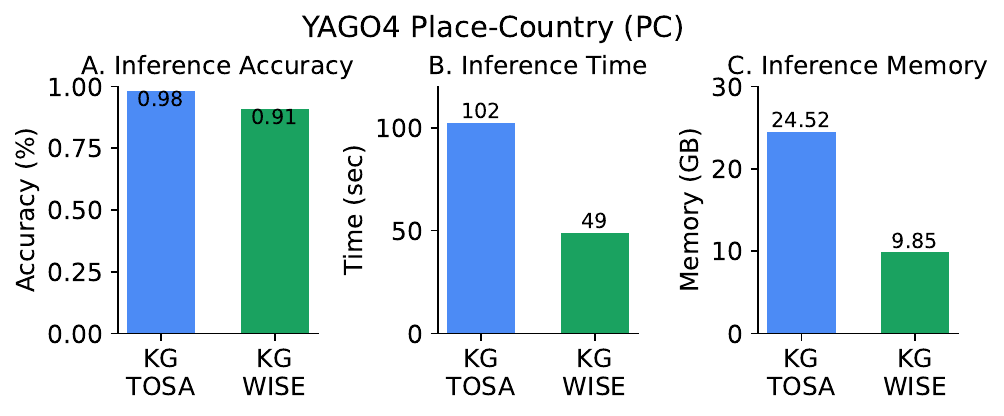}
         \label{fig:NC_KG-TOSA_WISE_YAGO}
     \end{subfigure}
     \vspace*{-4ex}
    \caption{{\TOSA} vs.\ {\sysName} on node classification tasks using three metrics: (A) Inference Accuracy (higher is better), (B) Inference Time (lower is better), and (C) Inference Memory (lower is better). Our LLM-guided subgraph extraction and query-aware inference substantially improve efficiency: {\sysName} reduces inference time by 41--52\% and memory usage by 60--84\% while preserving accuracy.}

    \label{fig:KG-TOSA_vs_KGWISE_NC}
    \vspace*{-4ex}
\end{figure}

\noindent\textbf{LLM-Guided Subgraphs for Training and Inference:}
We compare {\sysName} with {\TOSA} to quantify the effect of LLM-guided subgraph extraction. Both systems use subgraphs during training,  but {\TOSA} relies on fixed graph patterns, while {\sysName} selects semantically relevant relations using an LLM-guided query template. We evaluate both approaches across three NC tasks on DBLP, MAG, and YAGO4, using GraphSAINT as the backbone in all cases (Figure~\ref{fig:KG-TOSA_vs_KGWISE_NC}).\shorten

% \noindent\textit{\textbf{Training impact:}}
% Fixed graph patterns in {\TOSA} include many irrelevant triples, which increases subgraph size and training time. In contrast, {\sysName} filters the schema using semantic cues from the LLM, producing more compact training subgraphs. As a result, training time is reduced by 2--3$\times$ (e.g., DBLP: 26.9$\rightarrow$11.9 min, MAG: 59.3$\rightarrow$34.3 min, YAGO4: 57.9$\rightarrow$16.8 min). Training memory usage is also up to 78\% lower, with accuracy remaining comparable. 

\noindent\textit{\textbf{Training impact:}}
Fixed graph patterns in {\TOSA} include many irrelevant triples, which increases subgraph size and training time. In contrast, {\sysName} filters the schema using semantic cues from the LLM, producing more compact training subgraphs. As a result, training time is reduced by 2--3$\times$ (e.g., DBLP: 26.9$\rightarrow$11.9 min, MAG: 59.3$\rightarrow$34.3 min, YAGO4: 57.9$\rightarrow$16.8 min). Training memory usage is also up to 78\% lower, with accuracy remaining comparable.\shorten

\noindent\textit{\textbf{Inference impact:}}
During inference, {\TOSA} loads the full model into memory and performs a complete forward pass, while {\sysName} instantiates a compact model by loading only the embeddings and weights required for the extracted subgraph. As shown in Figure~\ref{fig:KG-TOSA_vs_KGWISE_NC}, this leads to 41--52\% faster inference and 60--84\% lower memory usage across all datasets. The gains are largest on dense or highly heterogeneous graphs such as MAG and YAGO4, where {\TOSA} loads many irrelevant node embeddings. In contrast, {\sysName} restricts computation to the semantically relevant neighborhood of the query. In the few cases where {\TOSA} achieves slightly higher accuracy, this comes from training on larger subgraphs, but it also causes up to 6$\times$ higher memory usage and nearly double the inference time. These results confirm that LLM-guided subgraph selection enables scalable, query-aware inference while preserving accuracy.

\begin{figure}[t]
\vspace*{-2ex}
     \centering
     % \hfill
     \begin{subfigure}
         \centering
         \includegraphics[width=0.35\textwidth]
         % {figures/Experiments/NC_DBLP/dblp_chunks.pdf}
         {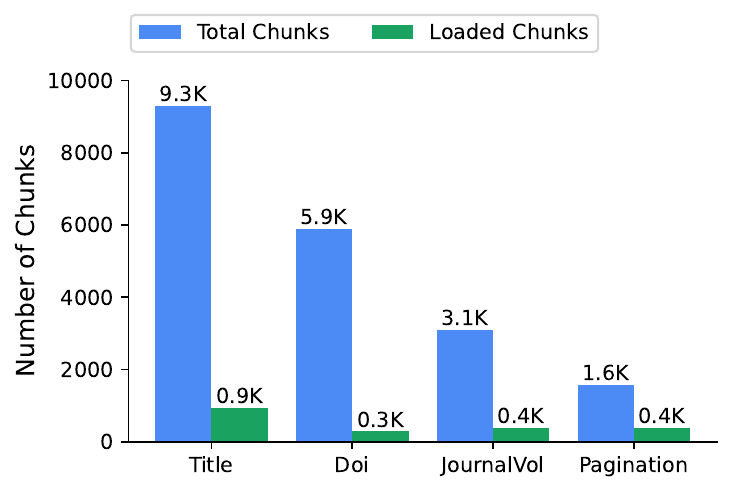}
         \label{fig:DBLP_chunks}
         % \vspace*{-3ex}         
     \end{subfigure}    
     \vspace*{-3ex}
    \caption{ The number of KV-store chunks loaded per node type by {\sysName} to answer the 1K target node inference query on the DBLP KG. {\sysName} loaded just 10\% of the KV-chunks, achieving significant savings in both loading time and memory.\shorten
    } 
\label{fig:Chunk_Eval}
  \vspace*{-4ex}
\end{figure}

\noindent\textbf{{\sysName} and KV-Chunks Access:}
The efficient chunking and indexing of neighbor node embeddings in {\sysName} enables the system to load only the embeddings of the nodes specified in the inference subgraph. This selective loading significantly reduces memory overhead and inference time.
Figure ~\ref{fig:Chunk_Eval} illustrates the number of loaded chunks per node type for the DBLP NC task using {\sysName}. The figure highlights the top 4 most frequent node types and shows the total number of chunks and the chunks loaded by {\sysName}, showcasing the effectiveness of {\sysName} in minimizing unnecessary data loading. Notably, only 10\% of the total chunks were required, demonstrating substantial computational and memory savings.
By efficiently querying the KV-store indexed chunks, {\sysName} achieves significant savings in embedding loading time and memory consumption resulting in a smaller embedding memory footprint in $\widetilde{M}$. Furthermore, indexing node embeddings within the KV-store facilitates scalable storage for large-scale KG embeddings while ensuring a lightweight GNN inference query complexity, making {\sysName} well-suited for high-throughput applications.\shorten

\begin{figure}[t]
\vspace*{-2ex}
     \centering
     % \hfill
     \begin{subfigure}
         \centering
         \includegraphics[width=0.48\textwidth]
         {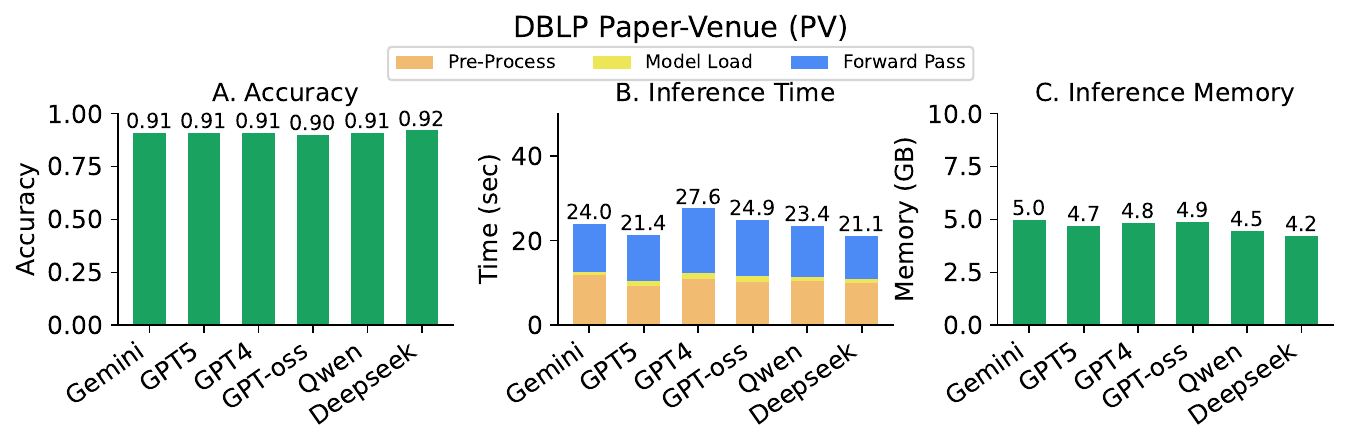}
         \label{fig:WISE_LLM_DBLP}
         \vspace*{-2ex}         
     \end{subfigure}
     % \hfill
     \begin{subfigure}
         \centering 
         \includegraphics[width=0.48\textwidth]{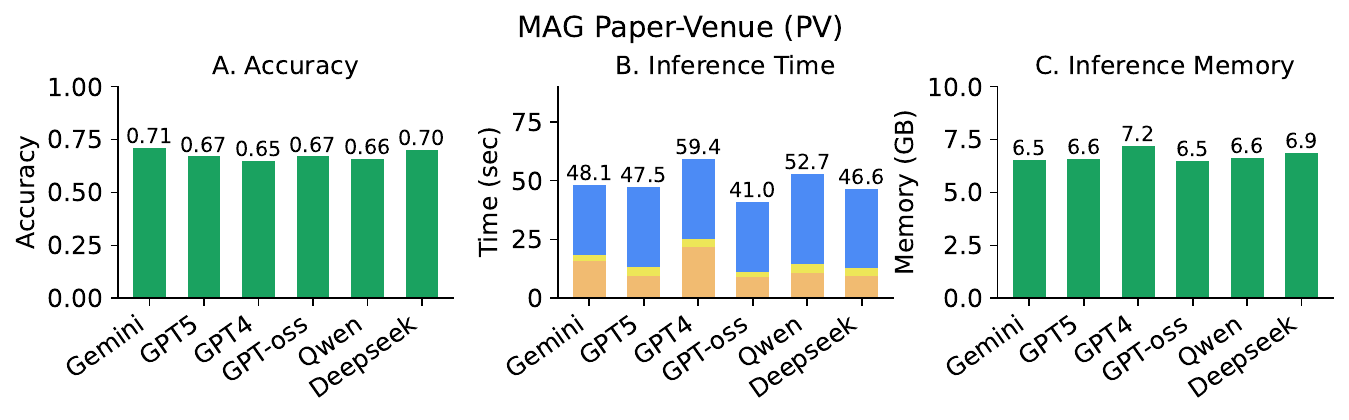}
         \label{fig:WISE_LLM_MAG}
         \vspace*{-2ex}         
     \end{subfigure}
     \begin{subfigure}
         \centering
         \includegraphics[width=0.48\textwidth]       {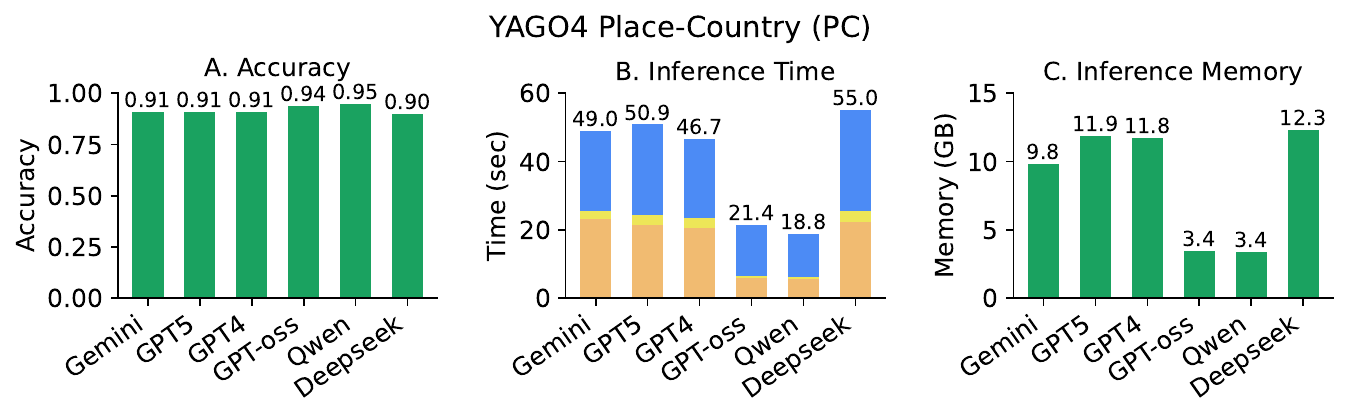}
         \label{fig:WISE_LLM_YAGO}
     \end{subfigure}
     \vspace*{-5ex}
    \caption{
Evaluation of different LLMs used to generate the query template $Q_T$ in {\sysName} across three KG tasks. 
Accuracy remains stable across all models, while inference time and memory vary due to differences in predicate selection. 
Open-weight LLMs (Qwen, GPT-oss) produce more compact subgraphs on heterogeneous KGs (e.g., YAGO4), yielding lower memory usage and faster inference compared to proprietary LLMs. 
}

    \label{fig:WISE_LLMs}
    \vspace*{-4ex}
\end{figure}

\noindent\textbf{LLM-Agnostic Template Generation and Inference Efficiency.}
{\sysName}'s prompt-engineering pipeline splits query generation into verifiable subtasks (Algorithm~\ref{alg:query_gen}), which stabilizes LLM behavior and prevents drift. This staged verification ensures that even lightweight open-weight models can generate $Q_T$ of comparable quality to proprietary models. As a result, {\sysName} generalizes across LLM types while enabling a tunable trade-off between subgraph compactness and resource usage.
Figure~\ref{fig:WISE_LLMs} compares {\sysName} across six LLMs. Accuracy remains stable on all datasets, confirming that inference quality does not depend on which LLM produces the template $Q_T$. The main differences arise in inference time and memory footprint, driven by how each $Q_T$ selects predicates during subgraph extraction.
On DBLP, all models behave similarly because the schema is shallow and low in heterogeneity. On MAG, the denser relation structure slightly reduces accuracy across all LLMs but runtime differences remain small. The largest contrast appears on YAGO4: open-weight models (Qwen, GPT-oss) select more compact predicate sets, yielding smaller subgraphs and reducing memory to $\sim$3.4\,GB, whereas proprietary models (Gemini, GPT-4, GPT-5) expand the neighborhood scope and push memory usage to 9–12\,GB.
These results show that “larger LLM” does not imply “better $Q_T$.” Compact predicate selection often leads to faster and cheaper inference without sacrificing accuracy.
% \vspace*{-4ex}

\begin{table}[!t]
\vspace*{1ex}
\centering
\setlength{\tabcolsep}{1pt} 
% \caption{The CO$_2$ emissions (in grams) and energy consumption (in Watt-hours) of {\sysName} and Graph-SAINT are compared. {\sysName} generates 60\% less CO$_2$ emissions and consumes 62\% less energy.}
\caption{CO$_2$ emissions and energy consumption of {\sysName} and Graph-SAINT. {\sysName} generates 60\% less CO$_2$ emissions and consumes 62\% less energy.}
\label{tbl_Carbon_Emission}
\vspace*{-2ex}
\small
\begin{tabular}{|c|c|c|}
\hline
\textbf{Metric} & \textbf{GraphSAINT} & \textbf{{\sysName}} \\
\hline
\textbf{Total Energy (Wh)}    & 1.19     & 0.42     \\
\hline
\textbf{Energy per Target Node (Wh)} &  1.2$\times 10^{-3}$&    0.4$\times 10^{-3}$  \\
\hline
\textbf{Total Emission (g CO$_2$)}  &  1.5$\times 10^{-1}$   & 0.53$\times 10^{-1}$  \\
\hline
\textbf{Emission per Target Node (g CO$_2$)} & 1.5$\times 10^{-4}$    & 0.53$\times 10^{-4}$ \\ 
\hline
\textbf{Relative CO$_2$ Efficiency}   & -    & 2.83x \\ 
\hline
\end{tabular}
\vspace*{-4ex}
\end{table}

% \begin{table}[]
% \begin{tabular}{lrr}
%                                      & \multicolumn{1}{c}{GraphSAINT} & \multicolumn{1}{c}{KGWISE}    \\ \cline{2-3} 
% \multicolumn{1}{l|}{Total Energy(kWh $\times 10^4$)}    & \multicolumn{1}{r|}{11.98}     & \multicolumn{1}{r|}{4.16}     \\
% \multicolumn{1}{l|}{Total Emission(Kg CO_2 $\times 10^4$)}  & \multicolumn{1}{r|}{0.011}     & \multicolumn{1}{r|}{0.00041}  \\
% \multicolumn{1}{l|}{Emission per TN(kWh $\times 10^4$)} & \multicolumn{1}{r|}{1.50}      & \multicolumn{1}{r|}{0.53}     \\
% \multicolumn{1}{l|}{Energy per TN(Kg CO_2 $\times 10^4$)}   & \multicolumn{1}{r|}{0.0015}    & \multicolumn{1}{r|}{0.000053} \\ \cline{2-3} 
% \end{tabular}
% \end{table}

% Total Energy (kWh $\times 10^4$)
% Total Emission (Kg CO_2 $\times 10^4$)
% Energy per TN (kWh $\times 10^4$)
% Emission per TN (Kg CO_2 $\times 10^4$)

%\noindent\textbf{Energy and Carbon Emissions.}
\subsection{Energy and Carbon Emissions.}
\vspace*{-1ex}
We measure the inference carbon footprint of {\sysName} and GraphSAINT using CodeCarbon~\cite{CodeCarbon}. CO$_2$ emissions are computed as $C \times E$, where $E$ is the measured energy consumption (CPU + memory) and $C$ is the carbon intensity of the electricity source recorded by CodeCarbon. Only CPU and memory energy are included; memory is estimated at 0.375\,W/GB and CPU usage is read from Intel RAPL.
Table~\ref{tbl_Carbon_Emission} reports the energy usage and CO$_2$ emissions when running inference on 1K DBLP target nodes. {\sysName} consumes 0.42~Wh compared to 1.19~Wh for GraphSAINT, a 62\% reduction. This translates to 3$\times$ lower energy per target node. The total CO$_2$ emissions drop from $1.5\times10^{-1}$\,g to $0.53\times10^{-1}$\,g, a 60\% reduction. The relative CO$_2$ efficiency of {\sysName} is therefore 2.83$\times$. These savings stem from avoiding full-model loading and reducing forward-pass computation, showing that {\sysName} improves system efficiency while also lowering environmental cost for better sustainability.\shorten

% \input{sections/RelatedWork.tex}
%\vspace*{4ex}
\section{Conclusion}
\label{sec:conclusion} 

In heterogeneous GNNs over large KGs, the majority of the model size comes from embeddings of non-target nodes rather than trainable weights. As our analysis shows, embeddings can account for more than 95\% of the total model size, while weights contribute only a small fraction. Existing systems must still load these embeddings in full at inference time, even when only a small task-specific subgraph is relevant. This results in unnecessary memory consumption and computation.
{\sysName} addresses this inefficiency by decoupling the GNN model into fine-grained components and enabling query-aware retrieval. An LLM-guided query template, generated once before training, is used to extract a compact task-relevant subgraph for each inference query. {\sysName} then instantiates a smaller model $\widetilde{M}$ by loading only the embeddings and parameters needed for that subgraph, avoiding full model materialization.
Our experiments on six real-world KGs show that this design significantly reduces inference memory usage and latency while maintaining or improving accuracy, as well as saving energy. These results demonstrate that partial model loading and query-aware adaptation are practical and effective strategies for scalable and sustainable GNN inference in large heterogeneous KGs.

% GNNs are effective for both homogeneous and heterogeneous graphs, but heterogeneous graphs such as KGs pose scalability challenges because their diverse node and edge types lead to large GNN models. Inference queries typically focus on a small subset of target nodes and their neighborhoods. However,  existing methods load and compute over the entire model and KG. Our empirical study reveals that this leads to the loading of irrelevant components, which highlights the need for query-specific model instantiation and partial loading.
% %
% This paper presents {\sysName}, a scalable system for efficient GNN inference on large KGs. {\sysName} (a) decomposes trained GNN models into fine-grained components stored in a key-value store, (b) loads only the components needed for each query, and (c) instantiates compact models over query-specific subgraphs extracted via LLM-guided semantic reasoning. {\sysName} utilizes LLMs to identify relevant predicates and features to generate SPARQL queries that extract subgraphs centered on target nodes. This process is optimized to enable fast and efficient query-aware model inference.
% %
% Experiments on six real-world KGs show that {\sysName} reduces inference latency by up to 90\%, memory usage by 80\%, and carbon footprint by 60\%, while maintaining or improving accuracy. {\sysName} advances sustainable AI by optimizing system design for efficient and eco-friendly GNN serving.

\balance
\bibliographystyle{IEEEtran}
\bibliography{refrences}
\onecolumn 
% \appendix
\section{Appendix}
\label{sec:appendix}
\subsection{The LLM-Guided Subgraph Extraction Prompts}
\label{appendix_prompts}
\vspace{2ex}
\textbf{ Suggest GNN Features Prompt}
\begin{boxE}
{
\label{prompt_suggest_feaatures}
\small
-You are an expert in machine learning feature selection, specifically for the GNN graph machine tasks.\\
-Think about information required to accurately \textbf{\texttt{<task>}}.\\
- Return a numbered list of items without explanation.\\
- Sort the list according to item importance.
}
\end{boxE}
\textbf{DBLP Example Output}
\begin{boxE}
{
\label{prompt_suggest_feaatures_example}
\small
1.  Keywords/Topics extracted from the publication's title and abstract\\
2.  Venues of publications cited by this publication\\
3.  Venues of previous publications by the authors of this publication\\
4.  Abstract text or embeddings of the publication\\
5.  Title text or embeddings of the publication\\
6.  Research domains/sub-fields associated with the publication's content\\
7.  Co-author network's historical venues\\
8.  Common entities (e.g., specific algorithms, datasets, tools) mentioned in the publication\\
9.  Average prestige/impact factor of venues where authors have previously published\\
10. Years of publication of cited works\\
11. Author's affiliation type or research focus\\
12. Publication year of the current paper
}
\end{boxE}
% \vspace{-2ex}
\textbf{Features to BGPs Mapping Prompt}
\begin{boxE}
{
\label{prompt_Frature_to_BGP}
\small
-You are an expert in machine learning feature selection for graph machine learning tasks.\\ 
- The following describes the \textbf{\texttt{<KG>}} knowledge graph schema, detailing the relationships between graph entities in a series of triples, one triple per line:\\
\textbf{\texttt{<KG-schema>}}\\
-Given the following list of key features, select the matching relations from the previous schema.\\
\textbf{\texttt{<suggested-features>}}\\
-Think carefully and refine your selected/matching items
  Return the top \textbf{\texttt{<K>}} matched schema triples sorted by importance.\\
-Output only one selected triple per line without any explanation.
}
\end{boxE}
\textbf{DBLP Schema BGP's and their stats \textbf{\texttt{<KG-schema>}} as features to help LLM decide the query's relevant BGP's}
\begin{boxE}
{
\label{prompt_suggest_feaatures_example}
\small
\textbf{subject\hspace{1.8cm}, predicate\hspace{0.2cm}, object , count}\\
schema:Publication , schema:title , str , 6008680\\
schema:Publication , dblp:yearOfEvent , purl:ResourceIdentifier , 11404719\\
schema:Publication , dblp:publishedInJournalVolume , str , 10256566\\
schema:Publication , schema:bibtexType , str , 6008680\\
schema:Publication , rdfs:label , str , 6008680\\
schema:Publication , schema:authoredBy , str , 5808242\\
schema:Publication , schema:publishedInSeries , str , 3057949\\
schema:Publication , schema:numberOfCreators , str , 6008680\\
schema:Publication , schema:publishedBy , schema:Publication , 3057103\\
schema:Publication , schema:doi , str , 4842535\\
...\\
}
\end{boxE}

% \vspace{-2ex}
\newpage
\subsection{SPARQL generation and refinement prompts}
\textbf{BGPs To SPARQL Prompt}
\begin{boxE}
{
\label{prompt_BGP_to_SAPRQL_refine}
\small
-You are an expert SPARQL query writer.\\
- Given the following triples list from the \textbf{\texttt{<KG>}} knowledge graph schema, write a SPARQL query to select the \textbf{\texttt{<VT>}} and its associated information given in the following triples list.\\
- The triples are directed; make sure to fulfill the direction and relation type.\\
- The query must return the union of sub-select statements in the form ?s ?p ?o.\\
- Each triple is Subject Entity - relation - Object Entity.\\
- Start with the \textbf{\texttt{<VT>}} node.\\
\textbf{\texttt{<BGP-List>}}\\
\textbf{\texttt{<SPARQL-Example>}}\\
-----------------Rules---------------------\\
1- Write nested select sub-queries and Union them.\\
2- In single-hop nested select, make sure to start the first BGP with the variable ?s.\\
3- In tow-hop or more nested select:\\
      3.1 Start the first BGP with the variable ?s, then use other variable names for next BGPs.\\
      3.2 Use the last connected entity as the subject, as shown in the previous example.\\
4- Generate only the SPARQL query without any explanation.\\
5- Make sure to use each given BGP triple.\\
6- add the BGP:  'Values ?s \textbf{\texttt{{<VT-List>}}}.' to the end of each sub query.\\
7- Refine all rules and the query syntax.
8- Do invent new relations i.e, dblp:authoredBy can not be dblp:Authored, But you can start with ?o instead of ?s.\\
Example:\\
      ?s a dblp:Publication.\\
      ?s dblp:authoredBy ?o.\\
      --------- Should Be ---------\\
      ?o a dblp:author.
      ?s dblp:authoredBy ?o.
}
\end{boxE}
\vspace{5ex}
\textbf{SPARQL Refine Prompt}
% \vspace{-1ex}
\begin{boxE}
{
\label{prompt_BGP_to_SAPRQL_refine}
\small
-You are an expert SPARQL query writer.
Given the following SPARQL query, re-write it to follow the following rules.\\
- Rule1: Keep the nested selects and their Union statements.\\
- Rule2: restructure the n-hop sub-select to choose the latest BGP subject and object and as the select items.\\
- Example: \{?s a prefix:x. 
?s prefix:y ?y. 
?y prefix:z ?z.\}\\
the latest BGP is ?y prefix:z ?z, then the select items must be: 1- ?y as ?s.  2- ?p.  3- ?z as ?o\\
-------- SPARQL Query ----------------\\
\textbf{\texttt{<sparql-query>}}\\
-Refine The Rules and Examples Carefully.\\
-Return only the Query; do not return any explanation.\\
\texttt{<Answer>}
}
\end{boxE}
\newpage
    \textbf{An example of LLM-Guided SPARQL Query generated for the DBLP-PV NC task.}
\begin{boxE}
{
\definecolor{DarkGreen}{RGB}{0,100,0}
\label{prompt_BGP_to_SAPRQL_refine}
\small
\textcolor{DarkGreen}{
PREFIX dblp: \texttt{<https://dblp.org/rdf/schema\#>}\\
PREFIX rdfs: \texttt{<http://www.w3.org/2000/01/rdf-schema\#>}}\\
\textcolor{blue}{SELECT ?s ?p ?o }\\
\textcolor{red}{FROM \texttt{<https://www.dblp.org>}\\
WHERE \{}\\
\{ SELECT ?s ?p ?o WHERE \{ ?s a dblp:Publication. ?s dblp:title ?o. BIND( "dblp:title" AS ?p). VALUES ?s \{\textbf{\texttt{<VT-List>}}\}. \}\}  \textbf{UNION}\\
\{ SELECT ?s ?p ?o WHERE \{ ?s a dblp:Publication. ?s dblp:yearOfEvent ?o. BIND( "dblp:yearOfEvent" AS ?p). VALUES ?s \{\textbf{\texttt{<VT-List>}}\}. \}\} \\\textbf{UNION}\\
\{ SELECT ?s ?p ?o WHERE \{ ?s a dblp:Publication. ?s dblp:publishedInJournalVolume ?o. BIND( "dblp:publishedInJournalVolume" AS ?p). VALUES    ?s \{\textbf{\texttt{<VT-List>}}\}. \}\} \\\textbf{UNION}\\
\{ SELECT  ?author ?p ?o WHERE \{ ?s a dblp:Publication. ?s dblp:authoredBy ?author.  ?author dblp:primaryAffiliation ?o. BIND( "dblp:primaryAffiliation" AS ?p). VALUES ?s \{\textbf{\texttt{<VT-List>}}\}. \}\} 
\\ \textbf{UNION}\\
\{ SELECT ?author ?p ?o WHERE \{ ?s a dblp:Publication. ?s dblp:authoredBy ?author. ?author rdfs:label ?o. BIND( "rdfs:label" AS ?p). VALUES ?s       \{\textbf{\texttt{<VT-List>}}\}. \}\} 
\\ \textbf{UNION}\\
\{ SELECT ?s ?p ?o WHERE \{ ?s a dblp:Publication. ?s dblp:publishedInSeries ?o. BIND( "dblp:publishedInSeries" AS ?p). VALUES ?s \{\textbf{\texttt{<VT-List>}}\}. \}\} 
\\ \textbf{UNION}\\
\{ SELECT ?s ?p ?o WHERE \{ ?s a dblp:Publication. ?s dblp:numberOfCreators ?o. BIND( "dblp:numberOfCreators" AS ?p). VALUES ?s \{\textbf{\texttt{<VT-List>}}\}. \}\} \\\textbf{UNION}\\
\{ SELECT ?s ?p ?o WHERE \{ ?s a dblp:Publication. ?s dblp:publishedBy ?o. BIND( "dblp:publishedBy" AS ?p). VALUES ?s \{\textbf{\texttt{<VT-List>}}\}. \}\} \\\textbf{UNION}\\
\{ SELECT ?s ?p ?o WHERE \{ ?s a dblp:Publication. ?s dblp:doi ?o. BIND( "dblp:doi" AS ?p). VALUES ?s \{\textbf{\texttt{<VT-List>}}\}. \}\} \\ \textcolor{red}{    \}}
}
\end{boxE}

\newpage
\subsection{{\sysName} performance while varying the inference query size}
% \vspace{-10ex}

\newpage
\subsection{{\sysName} training pipeline v.s. existing GNN training and inference accelerators}
\begin{figure*}[!h]
     \centering
     % \hfill
     \begin{subfigure}
         \centering
         \includegraphics[width=0.99\textwidth]
         {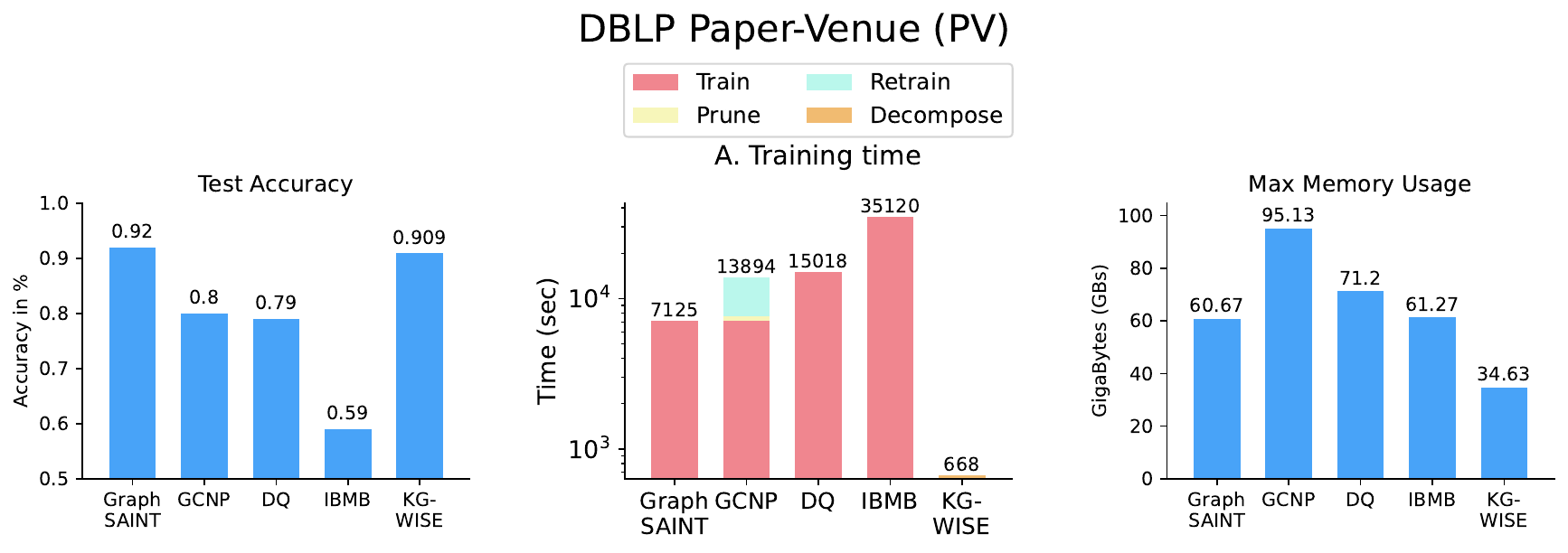}
         %{figures/Experiments/NC_MAG/MAG_INF_combined_plots.pdf}
         % \caption{DBLP}
         % \label{fig:DBLP-15M-RW}
         %\label{fig:MAG-42M-RW}
         \vspace*{-2.0ex}
         
     \end{subfigure}
     % \hfill
     
     \begin{subfigure}
         \centering 
         \includegraphics[width=0.99\textwidth]
         {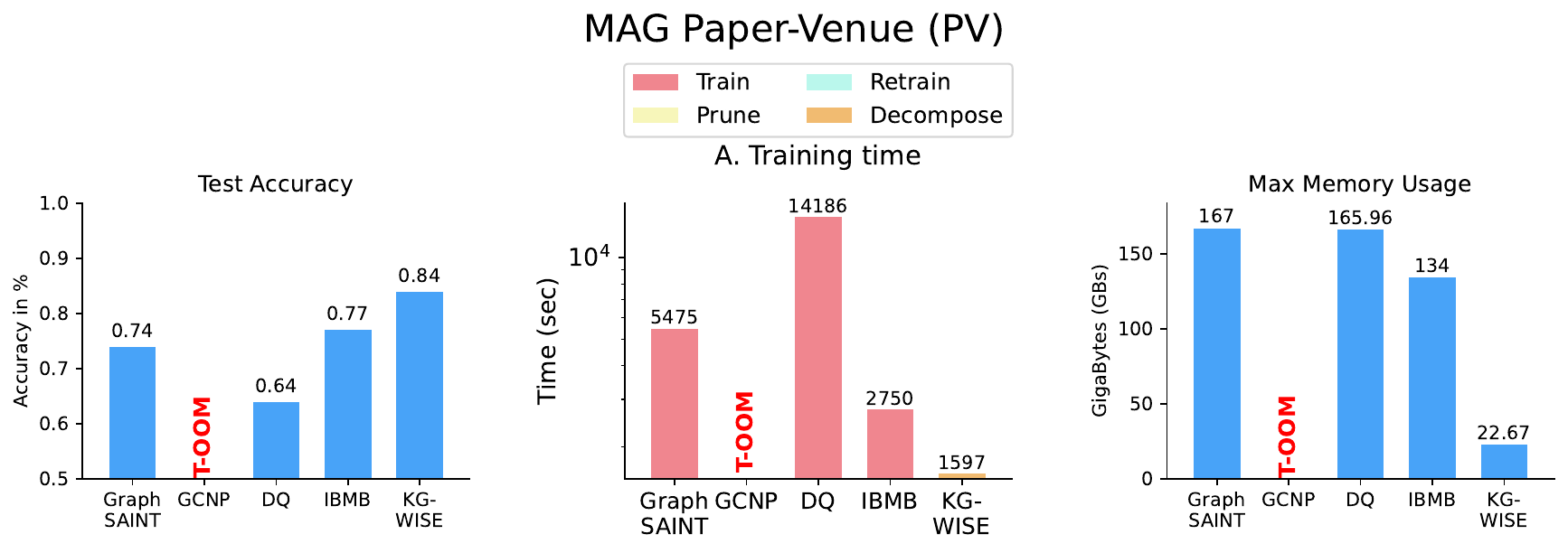}
         %{figures/Experiments/NC_DBLP/DBLP_INF_combined_plots.pdf}        
         % \caption{MAG}
         % \label{fig:MAG-42M-RW}
         %\label{fig:DBLP-15M-RW}
         \vspace*{-0ex}
     \end{subfigure}
     \begin{subfigure}
         \centering
         \includegraphics[width=0.99\textwidth]{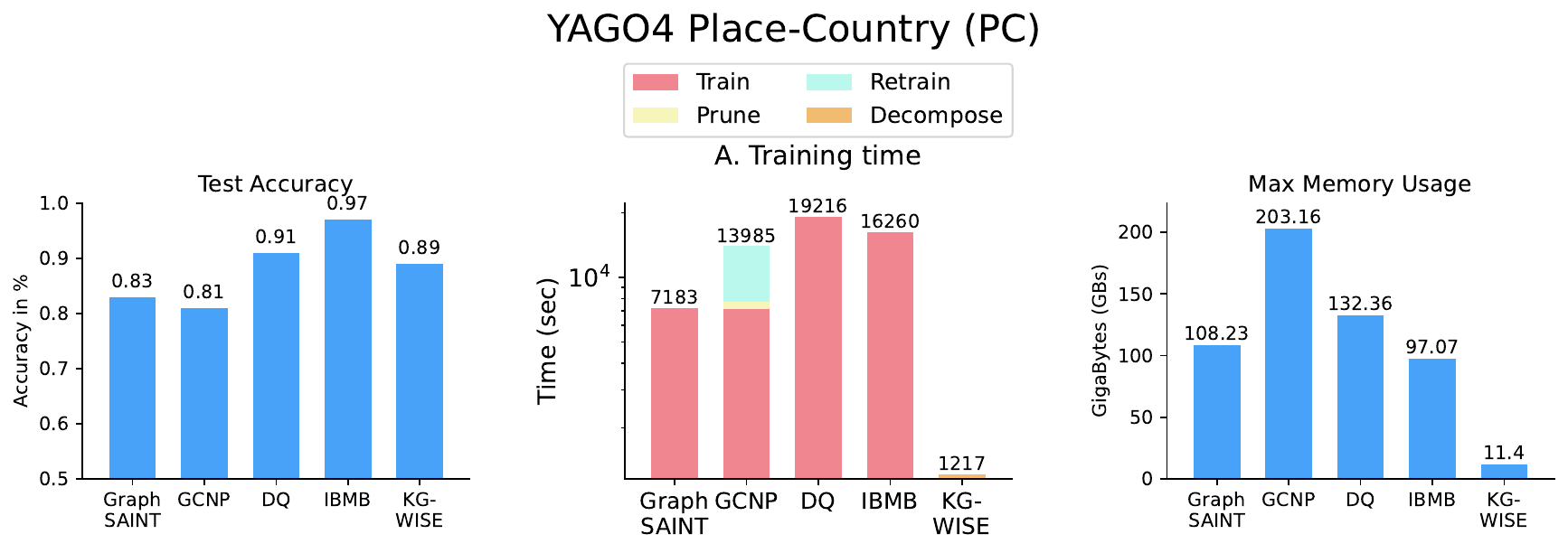}
         % \caption{YAGO-30M (Place-Country)}
         % \label{fig:YAGO-30M-RW}
     \end{subfigure}
     \vspace*{-2.0ex}
    \caption{ 
    Training metrics on NC tasks is measured using three metrics: (A) Test Accuracy (higher is better), (B) Training Time (lower is better), and (C) Max Memory Usage (lower is better). The top and middle sections show results for the Paper-Venue task on DBLP and MAG, respectively, while the bottom presents results for the Place-Country task on YAGO4.{\sysName} achieves comparable accuracy to SOTA methods and outperforms them with up to 6x faster training and 90\% memory reduction on YAGO4.} 
\label{fig:KGWISE_NC}
  \vspace*{-2ex}
\end{figure*}

\end{document}